\documentclass{article}

    \PassOptionsToPackage{numbers, compress}{natbib}
\usepackage[final]{neurips_2023}

\usepackage[utf8]{inputenc} % allow utf-8 input
\usepackage[T1]{fontenc}    % use 8-bit T1 fonts
\usepackage{hyperref}       % hyperlinks
\usepackage{url}            % simple URL typesetting
\usepackage{booktabs}       % professional-quality tables
\usepackage{amsfonts}       % blackboard math symbols
\usepackage{nicefrac}       % compact symbols for 1/2, etc.
\usepackage{microtype}      % microtypography
\usepackage{xcolor}         % colors
\usepackage{graphicx}
\usepackage{comment}
\usepackage{authblk}
\usepackage{amsmath}
\usepackage{wrapfig}
\usepackage{booktabs}
\usepackage[rightcaption]{sidecap}
\definecolor{VGG-19}{RGB}{174, 199, 232}
\definecolor{Inception-V3}{RGB}{34,120,181}
\definecolor{ResNet-50}{RGB}{47, 161, 72}
\definecolor{ViT-B}{RGB}{127, 127, 127}
\definecolor{Swin-B}{RGB}{158, 208, 138}
\definecolor{MLPMixer-B}{RGB}{214, 42, 40}
\definecolor{CORnet-S}{RGB}{247, 182, 209}
\definecolor{Harmonization}{RGB}{197, 177, 214}
\definecolor{ECOSET}{RGB}{199, 199, 199}
\definecolor{SimCLR}{RGB}{146, 104, 173}
\definecolor{MoCo-v3}{RGB}{196, 157, 149}
\definecolor{DINO}{RGB}{141, 87, 76}
\definecolor{MAE}{RGB}{216, 122, 178}
\definecolor{CLIP}{RGB}{245, 127, 32}
\definecolor{IPCL}{RGB}{188, 190, 50}
\definecolor{Noisy Student}{RGB}{246, 151, 151}
\definecolor{SWSL}{RGB}{252, 187, 120}

\definecolor{Blue}{RGB}{50,50,200}

\definecolor{Orange}{RGB}{255,127,80}

\title{SEVA: Leveraging sketches to evaluate alignment between human and machine visual abstraction}

\author[ ]{Kushin Mukherjee$^{1}$\footnote[1]{}, \ {Holly Huey$^{2}$\footnote[1]{}}, \ {Xuanchen Lu$^{2}$\footnote[1]{}}, \ {Yael Vinker$^{3}$}, \ {Rio Aguina-Kang$^{2}$}, \\{Ariel Shamir$^{4}$}, \ {Judith E. Fan$^{2,5}$}}
\affil[ ]{University of Wisconsin-Madison$^1$}
\affil[ ]{University of California, San Diego$^2$}
\affil[ ]{Tel-Aviv University$^3$}
\affil[ ]{Reichman University$^4$}
\affil[ ]{Stanford University$^5$}

\begin{document}

\footnotetext[1]{$^{/*}$Equal contribution}

\maketitle

\begin{abstract}

Sketching is a powerful tool for creating abstract images that are sparse but meaningful. Sketch understanding poses fundamental challenges for general-purpose vision algorithms because it requires robustness to the sparsity of sketches relative to natural visual inputs and because it demands tolerance for semantic ambiguity, as sketches can reliably evoke multiple meanings. While current vision algorithms have achieved high performance on a variety of visual tasks, it remains unclear to what extent they understand sketches in a human-like way. Here we introduce \texttt{SEVA}, a new benchmark dataset containing approximately 90K human-generated sketches of 128 object concepts produced under different time constraints, and thus systematically varying in sparsity. We evaluated a suite of state-of-the-art vision algorithms on their ability to correctly identify the target concept depicted in these sketches and to generate responses that are strongly aligned with human response patterns on the same sketch recognition task. We found that vision algorithms that better predicted human sketch recognition performance also better approximated human uncertainty about sketch meaning, but there remains a sizable gap between model and human response patterns. To explore the potential of models that emulate human visual abstraction in generative tasks, we conducted further evaluations of a recently developed sketch generation algorithm \cite{vinker2022clipasso} capable of generating sketches that vary in sparsity. We hope that public release of this dataset and evaluation protocol will catalyze progress towards algorithms with enhanced capacities for human-like visual abstraction.
\end{abstract}

% We conduct further evaluations of a recently developed algorithm \cite{vinker2022clipasso} capable of generating sketches that vary in sparsity to explore to what degree these sketches manifest signatures of human-generated sketches.

\section{Introduction}

%% Visual abstraction is key challenge for AI
%% Sketch generation and understanding as key case study in visual abstraction 
% \subsection{Sketches as paradigmatic case study in visual abstraction}

Abstraction is key to how humans understand the external world. 
Abstraction enables distillation of individual sensory experiences into compact latent representations that support learning of new concepts \cite{smith2002object, kemp2007learning, murphy2004big, goldstone2013concepts} and efficient communication about these concepts with others \cite{fan2020pragmatic, tessler2019language, hawkins2023visual, gentner2017analogy}.
For example, while no two roses are identical, people can rapidly infer what properties make a flower a \textit{rose} and not some other kind of flower from just a few examples \cite{xu2007word, lake2020people}, especially when these examples are selected to support such strong inferences \cite{gweon2010infants, shafto2014rational}.

\subsection{Human Visual Abstraction as Key Target for AI}
\begin{wrapfigure}{r}{0.35\textwidth}
    \centering
    \includegraphics[width= 0.35\textwidth]{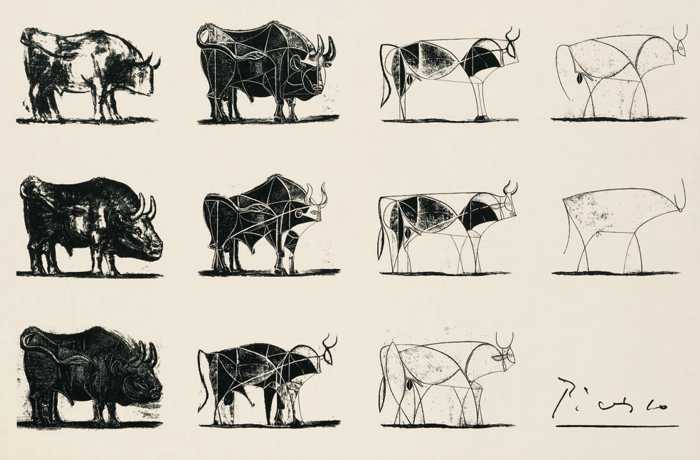}
    \caption{Pablo Picasso. \textit{The Bull}, 1945.}
    \label{fig:bulls}
\end{wrapfigure}

\textit{Visual abstraction} enables humans to express what they know about the visual world by creating external representations that highlight the information they judge to be most relevant in any given context---for instance, pictures that highlight the visual features that are diagnostic of a \textit{rose} in a botanical field guide \cite{fan2018common, fan2020pragmatic, viola2017pondering}. 
Critically, there are many different ways to depict even the same object---from a detailed illustration to a simple sketch. 
The Spanish artist Pablo Picasso famously demonstrated this point in \textit{The Bull} (1945), a series of 11 lithographs of bulls, each sparser than the last (Fig.~\ref{fig:bulls}).
While some of the drawings in this series look more realistic and others more stylized, all of these images remain evocative of a \textit{bull} (and perhaps other similar animals, such as a \textit{moose} or \textit{buffalo}) to most human viewers.

Drawing is one of the most accessible, enduring, and versatile techniques that humans in many cultures use to encode ideas and emotions in visual form \cite{hoffman2018dating, aubert2014pleistocene, gombrich1995story}. 
Even without special training, humans can robustly produce and understand simple line drawings or sketches of familiar visual concepts \cite{fan2018common, snodgrass1980standardized, jongejan2017quick}.
The ability to leverage drawings to understand and convey key aspects of the visual world emerges early in childhood \cite{luquet1927dessin, barrett1976symbolism, karmiloff1990constraints} and improves throughout development with children's expanding conceptual knowledge \cite{dillon2021rooms, long2021parallel, huey2022developmental}.
Moreover, failures to produce and recognize drawings of objects are associated with semantic dementia \cite{bozeat2003duck, rogers2007object}, suggesting links between a robust capacity for visual abstraction and the organization of semantic knowledge in the brain. 

In addition to drawings that represent objects and scenes, other abstract human-made visualizations (e.g., maps, diagrams, charts, graphs) serve important functions in many domains, including all branches of science and engineering \cite{tufte1986visual, tversky2000some, hegarty2011cognitive,chen2020foundations, card1999readings}. 
% Given the importance and ubiquity of these visualizations in modern life, developing computational models of human visual abstraction is thus key to developing AI systems that manifest stronger representational alignment between humans and machines. 
% Towards this longer-term objective, developing vision algorithms that achieve a robust and human-like understanding of hand-drawn images represents a key intermediate milestone.
Given the ubiquity and importance of such visualizations in modern life, developing computational models that achieve human-like understanding of freehand sketches is an important milestone.
Such computational models of \textit{human} visual abstraction stand to not only advance our understanding of human intelligence, but to also make AI systems more robust and general \cite{hendrycks2019benchmarking, madry2017towards,geirhos2018generalisation}. 
For example, prior work has found that incorporating principles based on the structure and function of the human visual system have led to vision models that are more robust (e.g., to adversarial attacks) \cite{battleday2020capturing, li2023recognizing, fel2022aligning}.
\vspace{-1 em}
\subsection{Desiderata for Evaluating Alignment Between Human and Machine Visual Abstraction}

% What are the core visual computations that support the ability to grasp the meaning of pictures, even when they vary across different levels of visual abstraction? 
% Semantic ambiguity as core feature of 

The past several years have seen remarkable progress in the development of increasingly performant general-purpose vision algorithms \cite{simonyan2014very, he2016deep, dosovitskiy2020image, radford2021learning}, with some of the most prominent algorithms also capable of emulating key aspects of how the primate brain encodes natural visual inputs \cite{yamins2014performance, kriegeskorte2015deep, zhuang2021unsupervised, konkle2022self}. 
% several architectural motifs inspired by the primate ventral visual stream \cite{gross1972visual,goodale1992separate,malach2002topography,hung2005fast}.
% Over the same time period, these artificial vision systems have also been steadily achieving higher performance on tasks related to visual abstraction, such as sketch categorization and sketch-based image retrieval \cite{eitz2012humans, eitz2012sketch, sangkloy2016sketchy, yu2016sketch, yu2017sketch, bhunia2020sketch, xu2022deep}, and been found to predict human error patterns on sketch recognition tasks to some degree \cite{fan2018common}. 
% In addition, the availability of larger models trained on ever vaster quantities of data have led to encouraging results on tasks involving images exhibiting different visual styles \cite{radford2021learning}, including sketch generation 
Over the same period, artificial vision systems have also been steadily achieving higher performance on tasks involving abstract visual inputs, including sketch categorization \cite{eitz2012humans, zhang2016sketchnet, ballester2016performance, yu2017sketch}, sketch segmentation \cite{li2018universal, yang2021sketchgnn}, sketch-based image/shape retrieval \cite{eitz2012sketch, sangkloy2016sketchy, yu2016sketch, song2017deep, bhunia2020sketch}, among others \cite{tripathi2020sketch, xu2022deep, lu2023learning, chowdhury2023can}. Moreover, these models have been found to predict human behavior on sketch recognition tasks to some degree \cite{fan2018common, fan2020pragmatic}. 
% \begin{wrapfigure}{r}{.4\textwidth}
% \centering
% \includegraphics[width=.4\textwidth]{neurips_figures/neurips_sketch_polysemy.pdf}
%     \caption{Sketches of a ``lion'' at different levels of sparsity. More abstract depictions invite greater semantic ambiguity.}
%     \label{fig:polysemy}
% \end{wrapfigure}
However, these otherwise high-performing vision algorithms struggle to simultaneously achieve robust understanding of visual inputs across multiple levels of abstraction \cite{baker2018abstract, singer2022photos, fan2020pragmatic}. 
Moreover, one study found that current vision models trained on natural images still fall short of the representational capabilities of the inferotemporal cortex, a key brain region supporting object categorization, in generalizing to new image distributions, including sketches \cite{bagus2022primate}.
% do not generalize as well to other image distributions, including sketches, as neurons in primate inferotemporal cortex, a key brain region supporting object categorization \cite{bagus2022primate}. 
Nevertheless, more recently developed models trained on substantially larger and varied datasets show promise on both tasks involving images with different visual styles \cite{radford2021learning, zhai2022scaling, schuhmann2021laion} and tasks that go beyond recognition, including sketch generation \cite{vinker2022clipasso, qiu2022emergent, wang2022sketchknitter} and sketch-guided image generation \cite{lu2018image, zhang2021sketch2model, wang20223d, voynov2023sketch, lin2023sketchfacenerf}.

At present, it remains unclear to what degree any state-of-the-art models achieve \textit{human-like} understanding of line drawings that vary in their degree of abstraction, much less the full range of abstract images that humans regularly engage with. 
Gaining further clarity on this question requires meeting two key challenges: \textit{first}, creating a dataset containing drawings of a wide variety of object concepts that also systematically vary in their degree of abstraction; and \textit{second}, developing evaluation protocols that can be used to estimate the degree to which any model emulates \textit{human-like} understanding of this suite of drawn images.

% clear metrics to estimate the degree to which any model emulates \textit{human-like} understanding of this suite of drawn images; and \textit{third}, conducting standardized evaluations of a broad suite of current vision algorithms that permit direct comparison between them and with humans. 

\textbf{Dataset.} 
% Meeting the first challenge requires going beyond existing sketch datasets, such as \textit{TU-Berlin} \cite{eitz2012sketch}, \textit{Quickdraw} \cite{jongejan2017quick}, and \textit{Sketchy} \cite{sangkloy2016sketchy}. 
% % \cite{gryaditskaya2019opensketch,manda2021cadsketchnet}
% While all of these datasets span a reasonably wide range of visual concepts (i.e., ranging from 125 concepts in \textit{Sketchy} to 345 in \textit{Quickdraw}), none of them systematically varied how detailed individual sketches could be, one of the most straightforward ways of inducing variation in semantic abstraction \cite{berger2013style, fan2020pragmatic, yang2021visual}. 
Meeting the first challenge requires going beyond existing sketch datasets \cite{eitz2012humans, sangkloy2016sketchy, jongejan2017quick, dey2019doodle, eitz2012sketch, li2018universal, yu2016sketch, song2017deep, Gao2020SketchyCOCO, wang2019spfusionnet, sarvadevabhatla2017object, li2014comparison, li2014shrec, li2017deeper, peng2019moment, zou2018sketchyscene, liu2020scenesketcher}. 
While most of these datasets span a reasonably wide range of visual concepts (i.e., ranging from 125 concepts in \textit{Sketchy} to 345 in \textit{Quickdraw}) and some of them contain fine-grained information (e.g., stroke information and photo-sketch pairing), none of them systematically varied how detailed individual sketches could be, one of the most straightforward ways of inducing variation in semantic abstraction \cite{berger2013style, fan2020pragmatic, yang2021visual}. 
Our paper addresses the gap by providing sketches with controlled levels of detail, while encompassing the variety and granularity present in existing datasets (\autoref{tab:dataset_summary}).

\textbf{Evaluation protocol.} 
% Meeting the second challenge requires going beyond measuring model performance alone. 
Meeting the second challenge requires going beyond simple accuracy-based model performance metrics alone.
Instead, it is critical to measure detailed patterns of human behavior on the same sketch understanding tasks to evaluate how well any model emulates these behavioral \textit{patterns}, following recent work in the computational neuroscience of vision \cite{rajalingham2018large, bear2021physion, peterson2019human}.

% \textbf{Evaluations.} Meeting the third challenge requires devising an approach to conducting fair comparisons between a wide variety of models that may vary in many ways (e.g., architectural commitments, how they were trained), with the goal of guiding future model development. 

%% Introducing SEVA: Sketch-based evaluations of visual abstraction 
\vspace{-1em}
\subsection{SEVA: A Novel Sketch Benchmark for Evaluating Visual Abstraction in Humans and Machines}
\vspace{-1em}
In recognition of the above desiderata, here we introduce \texttt{SEVA} (\textbf{S}ketch-based \textbf{E}valuations of \textbf{V}isual \textbf{A}bstraction), a new sketch dataset and benchmark for evaluating alignment between human and machine visual abstraction. 

\textbf{Dataset.} Our dataset contains approximately 90K human-generated sketches of a wide variety of visual objects that also systematically vary in their level of detail, and thus the variety of meanings they evoke. 
Each sketch is associated with one of 2,048 object instances belonging to one of 128 object categories selected from the \texttt{THINGS} dataset \cite{hebart2019things}. 
We achieved variation in sketch detail by imposing constraints on how much time humans ($N$=5,563 participants) had to produce each sketch (i.e., 4s, 8s, 16s, 32s). 

\textbf{Evaluation protocol.} Leveraging these human-generated sketches, we systematically evaluated how well a diverse suite of 17 state-of-the-art vision models generate classification responses that align with those produced by humans ($N$=1,709 participants) tasked with identifying the most appropriate concept label for each sketch. 
Of these participants, 579 participants also participated in the sketch production study but were not shown any of their own sketches during this study.
We evaluated human-model alignment using three different metrics: (1) top-1 classification accuracy, reflecting raw sketch recognition performance; (2) Shannon entropy of the response distribution, reflecting the degree of uncertainty about the target label; and (3) \textit{semantic neighbor preference}, reflecting the degree to which models and humans generated off-target responses that were semantically related to the target label. 

\begin{figure}[ht!]
    \centering
    \includegraphics[width=.85\textwidth]{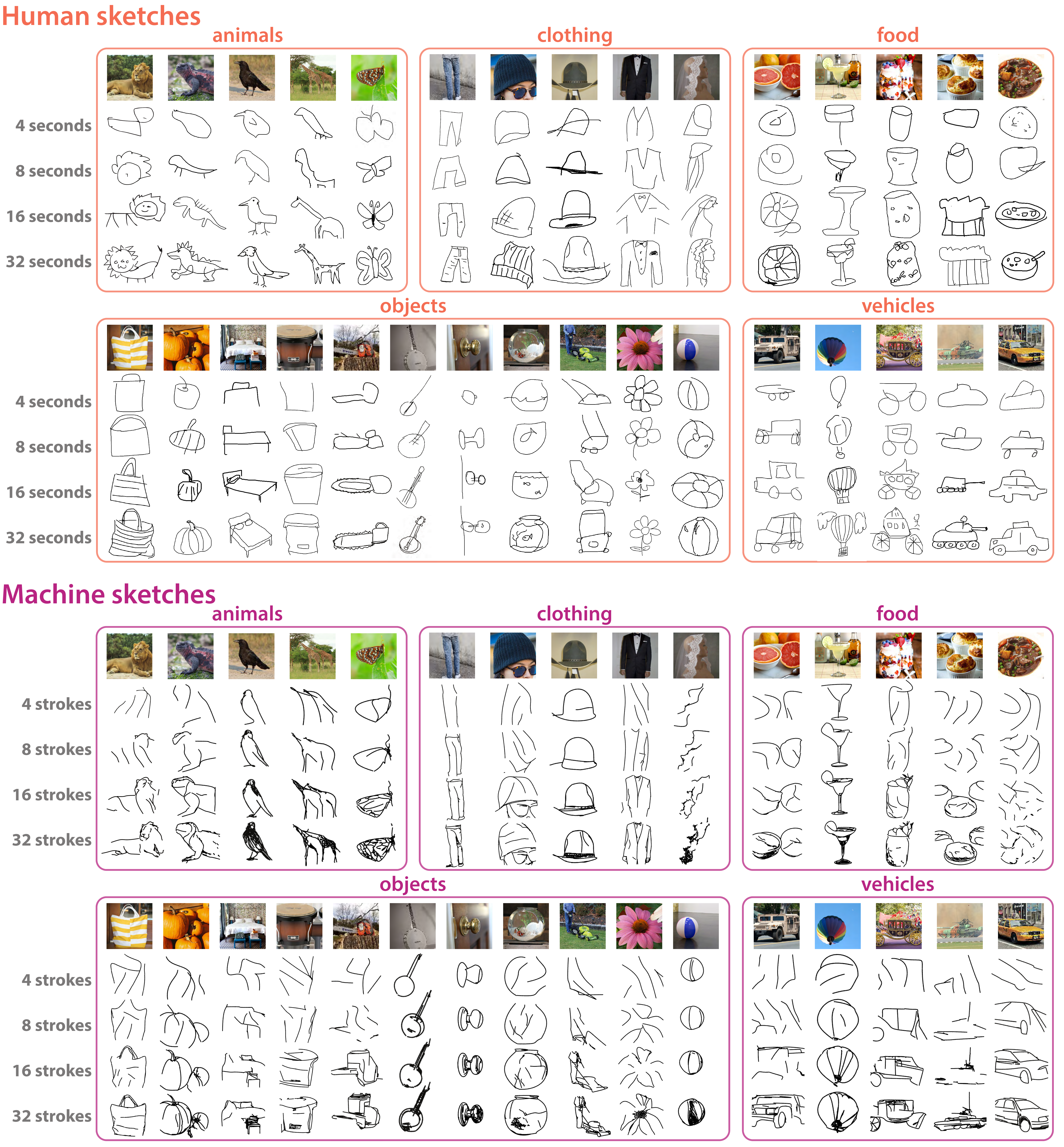}
    \caption{Humans and \texttt{CLIPasso} generated approximately 90K sketches under various production constraints.}
    \vspace{-1em}
    \label{fig:examplegallery}
\end{figure}

\textbf{Summary of key findings.} We found that sparser human sketches produced under more severe time pressure (e.g., 4 seconds) exhibited greater \textit{semantic ambiguity}---in other words, both humans and models assigned a greater variety of labels to them than to the more detailed sketches that took more time to make (e.g., 32 seconds).
Furthermore, we found that models that better predicted human sketch recognition performance also better approximated human uncertainty about sketch meaning, but none of the models came close to approximating human response patterns to human-generated sketches at any level of detail. 
To explore the potential of models that emulate human visual abstraction in generative tasks, we conducted further evaluations of \texttt{CLIPasso}, a recently developed sketch generation algorithm \cite{vinker2022clipasso} capable of generating sketches that vary in sparsity (measured by number of constituent strokes in a sketch). 
We discovered that the most detailed \texttt{CLIPasso}-generated sketches converged with human sketches of the same object concepts, as measured by the distribution of labels humans assigned to sketches made by both agent types; however, sparser \texttt{CLIPasso} sketches diverged from human sketches of the same object concepts, reflecting a gap between how \texttt{CLIPasso} and human participants attempted to preserve sketch meaning under more severe production constraints.

\vspace{-0.5em}
\section{Methods}
\vspace{-0.5em}
\subsection{Human Sketch Production}
\vspace{-0.5em}
A core contribution of this work is a new dataset containing human-generated sketches of a wide range of visual object concepts that also systematically span multiple levels of semantic abstraction. 
We created this dataset by crowdsourcing these sketches online, following prior work \cite{fan2018common, yu2016sketch, sangkloy2016sketchy, fan2020pragmatic, hawkins2023visual, jongejan2017quick}.
Each sketch in the dataset was recorded as a bitmap image as well as a collection of stroke coordinates, thus preserving the precise cursor movements a participant enacted to create the sketch.

\textbf{Participants.} 
5,563 participants (2,870 male; $M_{age}$ = 36.7 years) were recruited from Prolific and compensated $\$15.50$/hour for their participation.
Data from 104 of these sessions were excluded from subsequent analyses due to technical issues (e.g., images did not load). 
% We excluded data from 104 sessions due to technical glitches.
All participants provided informed consent in accordance with the UC San Diego IRB.
% our university's IRB (blinded for review). % 

\textbf{Object concepts.} We included 128 concrete real-world object categories (e.g., ``lion'', ``banjo'', ``car'') sourced from the \texttt{THINGS} dataset. 
We used the \texttt{THINGS} dataset \cite{hebart2019things, hebart2023things} because it is a well validated set of concrete, real-world visual object categories designed to support interoperability among large-scale studies in human visual cognition and cognitive neuroscience.
% Building on a prior study investigating human sketch production at different levels of abstraction \cite{yang2021visual}, we sampled concepts from the entire library of 1,854 concepts based spanning four main axes of variation: familiarity, artificiality, animacy, and size. 
For each of these 128 object concepts, we randomly sampled 16 object instances represented by color photographs, which served to visually ground the human sketch production task. 
As such, each sketch in our dataset is uniquely associated with one of these 2,048 object instances, and our final sample size was determined by our predefined goal of obtaining at least 10 human sketches of each of these instances. 
% color photographs furnishing a final set of 2,048 instance photographs, which served as referents for the human sketch production task.
% 2,048 images
% \subsection{Human sketch production Task}

\begin{table}[htp!]
\resizebox{\textwidth}{!}{%
\begin{tabular}{@{}llllll@{}}
\toprule
\textbf{Dataset Name} & \textbf{Dataset Contents} & \textbf{\# Classes} & \textbf{Stroke Info?} & \textbf{Photo Cue?} & \textbf{Abstraction?} \\ \midrule
TU-Berlin~\cite{eitz2012humans}   & 20K sketches & 250 & \checkmark  & & \\
QuickDraw~\cite{jongejan2017quick} & 50M sketches & 345  & \checkmark  & & \\
QuickDrawExtended~\cite{dey2019doodle}  & 330K sketches, 204K photos & 110   &  & & \\
SPG ~\cite{li2018universal} & 20K sketches w/ stroke grouping & 25 & \checkmark & &\\
SBSR~\cite{eitz2012sketch} & 1.8K sketches, 1.8K 3D models & 161  &  &  &\\
QMUL~\cite{yu2016sketch, song2017deep} & 1.3K sketches, 1.3K photos  & 3  &    & \checkmark & \\
Sketchy~\cite{sangkloy2016sketchy}     & 75K sketches, 12K photos  & 125  &  \checkmark &  \checkmark & \\
SketchyCOCO~\cite{Gao2020SketchyCOCO} &  14K sketches, 14K photos  & 17  & \checkmark  &\checkmark & \\
\textbf{SEVA} & \textbf{90K sketches, 2048 photos} & \textbf{128} & \checkmark  & \checkmark & \checkmark  \\ \bottomrule
\end{tabular}%
}
\vspace{.5em}
\caption{Comparison between \texttt{SEVA} and prior sketch datasets.}
\vspace{-2em}
\label{tab:dataset_summary}
\end{table}
% DomainNet: ClipArt, infographics, paintings, QuickDraw, real photos, professional sketches

\textbf{Sketch production task.} In each session, participants produced sketches of 16 different object categories, randomly sampled from the full set of 128 object categories.
On each trial, they were cued with a color photograph (500px $\times$ 500px) of an object paired with its concept label. 
Each participant was randomly assigned to one of four conditions, defined by the maximum amount of time participants could take to produce their sketches: 4 seconds, 8 seconds, 16 seconds, or 32 seconds (Fig.~\ref{fig:examplegallery}, \textit{left}). 
Such random assignment of participants to condition ensures that estimates of differences between conditions will not, in expectation, be biased by individual differences in sketching behavior.
% \edited{We chose a between-subjects randomization, because we predicted that it would be easier for participants to produce sketches within the same time constraint across all 16 sketches rather than adjusting to new a time constraint for each sketch.}
Participants drew on a digital drawing canvas (500px $\times$ 500px) using whatever input device they already had available (e.g., mouse, stylus) and were able to undo their most recent stroke or completely clear their canvas if needed. 
They were encouraged to make their drawings as recognizable as they could at the concept level and to use the photograph only to remind them of what individual objects belonging to that category generally look like.  
A countdown timer indicated how many seconds they had left to produce their drawing. 
Each trial ended either when time ran out or when the participant indicated that they wished to continue to the next trial, but participants were encouraged to use the full time available to produce as recognizable of a drawing as they could.
At the beginning of the session, participants were explicitly instructed not to include any background context (e.g., grass in a drawing of a ``horse''), arrows, or text.
Participants also completed one practice trial (that we did not include in analyses) at the beginning of the session to familiarize themselves with the drawing interface.
Our final dataset contains 89,797 sketches after filtering out invalid sketches (e.g., blank canvases).

\subsection{Human Sketch Understanding}
\vspace{-1em}
A key component of any evaluation of how well current vision algorithms emulate human visual abstraction is measurement of human behavior in tasks relying on visual abstraction. 
Here we focus on characterizing what meanings humans extract from the collected sketches, providing the basis for our subsequent empirical evaluation of how well any state-of-the-art vision model approximates human response patterns when presented with the same sketches.  

% In order to benchmark vision algorithms on their ability to understand the semantic content represented in sketches, we aimed to establish recognizability baselines for each sketch by acquiring dense soft-label human judgements \cite{collins2022eliciting,collins2023human, peterson2019human} for a subset of sketches produced in the previous experiment. 

\vspace{1em}
\textbf{Participants.}
1,709 participants (776 male; $M_{age}$ = 39.2 years) were recruited from Prolific and compensated $\$15.50$/hour for their participation. % were recruited to make recognition judgments on the 8,192 sketches.
Data from 21 of these sessions were excluded from subsequent analyses due to technical issues.
% We excluded 28 data sessions from participants who experienced technical difficulties. 
Our predefined criterion for stopping data collection was acquisition of at least 12 recognition judgments for each sketch.

\textbf{Sketch recognition task.}
In each session, participants provided labels for 64 sketches randomly sampled from a fixed set of 8,192 sketches, approximately 10\% of the full human sketch dataset. 
The specific set of 8,192 sketches included in this experiment was determined by randomly sampling one sketch cued by each object instance from each drawing-time condition (i.e., 16 instances/concept $\times$ 128 categories $\times$ 4 drawing-time conditions = 8,192). 
On each trial, participants were presented with a single sketch (300px $\times$ 300px) and a text field where they could provide their best guess concerning the concept the sketch was intended to convey. 
As soon as they began typing, a drop-down menu appeared with suggested word completions.
This drop-down menu contained the entire set of 1,854 labels in the \texttt{THINGS} dataset and only responses that matched one of these 1,854 labels were accepted. 
Because many words have multiple meanings, the labels contained in this dropdown menu were also accompanied by disambiguating text (e.g., to distinguish \textit{mouse (animal)} from \textit{mouse (computer)}). 
If participants were unsure of which label best applied to the sketch, they were encouraged to provide additional guesses (up to 5 per sketch). 
Collecting multiple labels on each drawing trial was important because it enabled us to more thoroughly sample the distribution of meanings that each sketch evoked for human participants (i.e., which labels came to mind and how often they did so). 
At the beginning of the session, participants completed a practice trial (that we did not include in analyses) to familiarize themselves with the labeling interface.

\subsection{Machine Sketch Understanding}

We propose a generic protocol for evaluating machine sketch understanding that can be applied to any vision algorithm using our sketch dataset. 
In this paper, we conduct evaluations of a wide range of state-of-the-art vision models with the goal of demonstrating the feasibility of our protocol and guiding future model development. 

\textbf{Model Suite.} Specifically, we evaluated 17 vision models spanning a wide range of architectures and training methods (Table \ref{tab:models}), all of which have been demonstrated to achieve high performance on object recognition on standard datasets, such as ImageNet \cite{deng2009imagenet}. We also made sure to include variants of standard ConvNet and Transformer models that have gained traction within the field of computational cognitive neuroscience for their potential to close the gap between biological and artificial vision \cite{konkle2021beyond,fel2022aligning,kubilius2019brain,mehrer2021ecologically}.

\begin{table}[htp!]
\centering

\resizebox{.9\textwidth}{!}{%
\begin{tabular}{@{}llll@{}}
\toprule
\textbf{Model}  & \textbf{Architecture} & \textbf{Training Paradigm} & \textbf{Dataset}\\ 
\midrule
\textbf{\textcolor{VGG-19}{VGG-19}}~\cite{simonyan2014very} & VGG-19 & supervised & ImageNet \\ 
\textbf{\textcolor{Inception-V3}{Inception-V3}}~\cite{szegedy2016rethinking} & Inception-V3 & supervised & ImageNet\\
\textbf{\textcolor{ResNet-50}{ResNet-50}}~\cite{he2016deep} &  ResNet-50  & supervised & ImageNet\\
\textbf{\textcolor{ViT-B}{ViT-B}}~\cite{dosovitskiy2020image} & ViT-B & supervised & ImageNet\\
\textbf{\textcolor{Swin-B}{Swin-B}}~\cite{liu2021swin} & Swin-B & supervised & ImageNet\\
\textbf{\textcolor{MLPMixer-B}{MLPMixer-B}}~\cite{tolstikhin2021mlp} & MLPMixer-B & supervised & ImageNet\\
\textbf{\textcolor{CORnet-S}{CORnet-S}}~\cite{kubilius2019brain} & CORnet-S & supervised & ImageNet\\
\textbf{\textcolor{Harmonization}{Harmonization}}~\cite{fel2022aligning} & ViT-B & supervised & ImageNet + Human Feature Importance~\cite{fel2022aligning}\\
\textbf{\textcolor{ECOSET}{ECOSET}}~\cite{mehrer2021ecologically} & ResNet-50 & supervised & ECOSET~\cite{mehrer2021ecologically}\\
% \textbf{\textcolor{MoCo-v2}{MoCo-v2}}~\cite{chen2020improved} & ResNet-50 & self-supervised \\
\textbf{\textcolor{SimCLR}{SimCLR}}~\cite{chen2020simple} & ResNet-50 & self-supervised & ImageNet\\
\textbf{\textcolor{MoCo-v3}{MoCo-v3}}~\cite{chen2021empirical} & ViT-B & self-supervised & ImageNet\\
\textbf{\textcolor{DINO}{DINO}}~\cite{caron2021emerging} & ViT-B & self-supervised & ImageNet\\
% \textbf{\textcolor{DINO}{DINO}}~\cite{caron2021emerging} & ResNet-50 & self-supervised & ImageNet\\
\textbf{\textcolor{MAE}{MAE}}~\cite{he2022masked} & ViT-B & self-supervised & ImageNet\\
\textbf{\textcolor{CLIP}{CLIP}}~\cite{radford2021learning} & ViT-B & self-supervised & WebImageText~\cite{radford2021learning}\\
\textbf{\textcolor{IPCL}{IPCL}}~\cite{konkle2021beyond} & AlexNet & self-supervised & ImageNet\\
\textbf{\textcolor{Noisy Student}{Noisy Student}}~\cite{xie2020self} & EfficientNet-b4 & semi-supervised & ImageNet + JFT~\cite{sun2017revisiting}\\
\textbf{\textcolor{SWSL}{SWSL}}~\cite{yalniz2019billion} &  ResNet-50  & semi-supervised & ImageNet + YFCC-100M~\cite{thomee2016yfcc100m} + IG-1B-Targeted~\cite{yalniz2019billion}\\
\bottomrule 
\end{tabular}%
}
\vspace{1em}
\caption{Model suite annotated by backbone architecture, training paradigm, and training dataset.}
\vspace{-1em}
\label{tab:models}
\end{table}

% We performed all our computations on latent features extracted from the deepest (non-fully-connected) layers of these models, which amounted to extracting activation patterns at either the model's final convolution or attention layer.

\textbf{Evaluation Protocol.}
The goal of our evaluation protocol was to measure how well any of these vision models approximated human sketch recognition behavior when presented with the same sketches. 
Because these models contain different latent representations of widely varying dimensionalities, we measured machine sketch-recognition behavior by extracting activation patterns from each model's final convolutional or attention block and training linear classification-based readouts on these activation patterns. 
That is, for each model we independently fit 1,854-way logistic regression classifiers using 5-fold stratified cross-validation to predict the ``ground-truth'' concept label associated with each sketch.
\footnote{Because these classification-based readouts were only trained on sketches of 128 object concepts subsetted from the \texttt{THINGS} dataset, the probabilities assigned to the remaining 1,726 labels were set to zero during training. As such, none of these models produced any of these other labels at test time, but humans in the sketch recognition experiment \textit{could} select these other labels. This difference between how sketch recognition behavior was elicited from humans and models led us to focus on relative measures of performance when evaluating human-model alignment.} 

% That is, for each model we independently fit 128-way logistic regression classifiers using 5-fold stratified cross-validation to predict the ``ground-truth'' category label from the feature embeddings of each sketch.
% \footnote{Because these classification-based readouts were only trained on sketches of 128 object concepts subsetted from the \texttt{THINGS} dataset, they never predicted any of the remaining 1,726 category labels. But humans in the sketch recognition experiment \textit{could} may select any label from the 1,854 labels. This difference between how sketch recognition behavior was elicited from humans and models led us to focus on relative measures of performance when evaluating human-model alignment.} 

These predicted labels were aggregated across the 16 sketches of the same concept (e.g., \textit{lion}) and from the same drawing-time condition (e.g., 4 seconds) to yield a model's response distribution for that \textit{type} of sketch.
The top-1 classification accuracy was determined by computing the relative frequency of the ``ground-truth'' concept label in this response distribution. 
We also computed the Shannon entropy of this response distribution to estimate the degree of semantic ambiguity exhibited by this type of sketch.
Further, we derived a measure of the degree to which even the \textit{non-}ground-truth labels generated by each model were semantically related to the ground-truth label, which we term the \textit{semantic neighbor preference} score.
This semantic neighbor preference score falls in the range $[0,1]$ and is highest when labels that are more semantically related to the ground-truth label appear more frequently than more semantically distant labels, is close to 0.5 when labels appear with uniform probability, and is minimized when labels that are more semantically distant appear most frequently. 

To compare model classification outputs with human responses, we derived an analogous response distribution from the human labels obtained in the human sketch recognition experiment. 
That is, we aggregated all labels assigned by human participants to all 16 sketches of the same concept (e.g., \textit{lion}) and from the same drawing-time condition (e.g., 4 seconds) to construct a response distribution for each type of sketch. 
We then computed the same three metrics above (i.e., top-1 classification accuracy, response entropy, semantic neighbor preference) using the human response distributions.

\subsection{Machine Sketch Production}

To explore the potential of models that emulate human visual abstraction in generative tasks, we also include evaluations of \texttt{CLIPasso}, a recently developed sketch generation algorithm \cite{vinker2022clipasso} capable of generating sketches that vary in sparsity.

\textbf{Generating machine sketches.} Specifically, we leveraged \texttt{CLIPasso} to generate 8,192 sketches conditioned on the same 2,048 object instances we used in the human sketch production experiment such that each sketch was constrained to consist of either 4, 8, 16, or 32 pen strokes (Fig.~\ref{fig:examplegallery}, \textit{right}). 
\texttt{CLIPasso} generates sketches by optimizing the parameters of a set of curves (i.e., start/end points; control points), each representing a single pen stroke, to be similar to target image.
This optimization is guided by a pretrained implementation of CLIP \cite{radford2021learning}, a large model trained using contrastive learning on vast quantities of text-image pairs.
Similarity to the target image is defined based on the distance between CLIP's embedding of the target image and its embedding of the sketch, where these embeddings reflect combinations of feature activations from multiple intermediate layers of CLIP. 

\textbf{Measuring human understanding of machine sketches.} We evaluated human recognition performance on these \texttt{CLIPasso}-generated sketches by recruiting 1,481 participants (730 male; $M_{age}$ = 41.05 years) on Prolific to complete the same sketch recognition task described earlier.
Data from 7 of these sessions were excluded from subsequent analyses due to technical issues.

% We excluded 28 participants who experienced technical difficulties.

% in intermediate layer activations on both the input image and the sketch.
% The abstraction degree is controlled by varying the number of strokes. 
% To investigate this observation, we utilize CLIPasso \cite{vinker2022clipasso}, a recently introduced method that generates sketches of objects at various abstraction levels.
% In CLIPasso, a sketch is represented as a set of parametric curves (strokes), whose parameters are optimized to depict a given image.
% The optimization is guided by a pretrained CLIP model, leveraging its powerful semantic and visual priors. 
% A CLIP-based loss is defined between the input image and the rasterized sketch, based on the distance between CLIP's latent embedding and its intermediate layer activations on both the input image and the sketch.
% The abstraction degree is controlled by varying the number of strokes. 

% We used CLIPasso to generate 8,192 sketches — one sketch for each of the 16 photographs in each of the 128 concept categories at 4 levels of abstraction (16 $\times$ 128 $\times$ 4).
% Because CLIPasso varies detail by modulating the number of strokes in a given sketch, to mirror the human sketch production task as closely as possible, we generated sketches using 4, 8, 16, and 32 strokes.

% \subsection{Statistical Analyses} Throughout this paper, we estimate the impact of our manipulation of drawing time and differences between models by fitting mixed-effects regression models, with random intercepts for different object concepts. 

\begin{wrapfigure}{r}{0.5\textwidth}
    \centering
    \includegraphics[width=.5\textwidth]{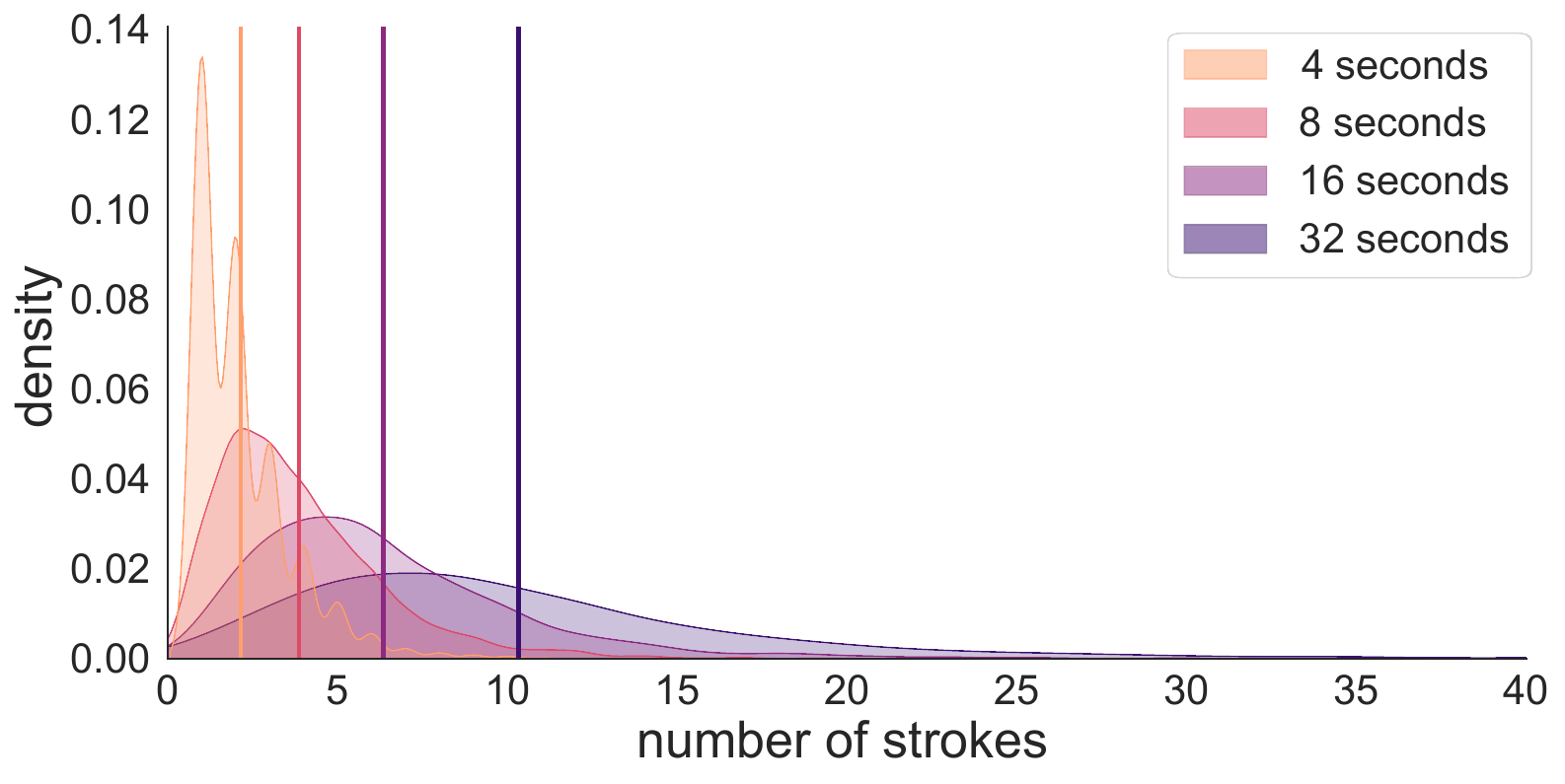}
    \caption{Distributions of number of strokes for each drawing time condition in the human sketch production task. Vertical lines indicate means.}
    \label{fig:dd_v_numstrokes}
    \vspace{-1em}
\end{wrapfigure}

\section{Results}
% \subsection{Benchmarking machine sketch understanding}

% Throughout, we estimate the impact of our manipulation of drawing time and differences between models by fitting mixed-effects regression models, with random intercepts for different object concepts. 

\textbf{Humans produce sparser sketches under stronger time constraints.}
We first sought to validate the effect of manipulating the maximum time that human participants had to draw on how detailed their sketches were. 
We estimated how detailed a sketch was by counting the number of strokes it contained (Fig.~\ref{fig:dd_v_numstrokes}) and then fit a mixed-effects linear regression model predicting the number of strokes as a function of drawing-time condition (i.e., 4s, 8s, 16s, 32s), with random intercepts for object concept.
We found that drawings produced under the 4s limit contained the fewest strokes on average, whereas those produced under the 32s limit contained the greatest number of strokes ($\beta=.29$, $SE=4.95 \times10^{-3}$, $p<.001$).
These results confirm that restricting the amount of time human participants had to produce their sketches led to systematic differences in how detailed their sketches were.
% These differences are critical for measuring human and model sensitivity and robustness to variable visual depictions of the same underlying object concept.

% % \vspace{-1em}
% \begin{figure*}
%     \centering
% \includegraphics[width=1\linewidth]{figures/VAB_polysemy_no_mocov2.pdf}
%     % \vspace{-1em}
%     \caption{\textit{Left:} Example of sketch polysemy at different levels of abstraction. \textit{Middle:} Proportion of variance explained in top-1 human recognition accuracy by model top-1 accuracy. \textit{Right:} Proportion of variance explained in human response distribution entropy by model class probability entropy. Inset scatterplots show the distribution of entropy values between human judgements and CLIP.}
%     \vspace{-1em}
%     \label{fig:VAB_polysemy}
% \end{figure*}
% % \subsubsection{Varying stroke complexity and draw duration  moderates perceived semantics in sketches for both humans and machines.}

\begin{table}[t!!]
    \centering
    \resizebox{.9\textwidth}{!}{
    \begin{tabular}{l|l|l||l|l||l|l}
    \toprule
    time & accuracy$_{mean}$ & accuracy$_{sem}$& entropy$_{mean}$ & entropy$_{sem}$& SNP$_{mean}$ & SNP$_{sem}$\\
    \toprule
        \textbf{4 seconds} & .031 & .003 & 1.958 & .011 & .628  & .008\\
        \textbf{8 seconds} & .082 & .004 & 1.814 & .013 & .701 & .009\\
        \textbf{16 seconds} & .139 & .006 & 1.690 & .014 & .750 & .010\\
        \textbf{32 seconds} & .199 & .007 & 1.555 & .015 & .787 & .010\\
        \bottomrule
      
    \end{tabular}}\\\
    \caption{Human sketch understanding under each draw duration constraint. Columns represent means and standard errors of the mean for top-1 accuracy, response entropy, and semantic neighbour preference (SNP).}
    \vspace{-2em}
    \label{tab:human_performance}
\end{table}

\begin{wrapfigure}{r}{.35\textwidth}    
    \centering
    \includegraphics[width=.35\textwidth]{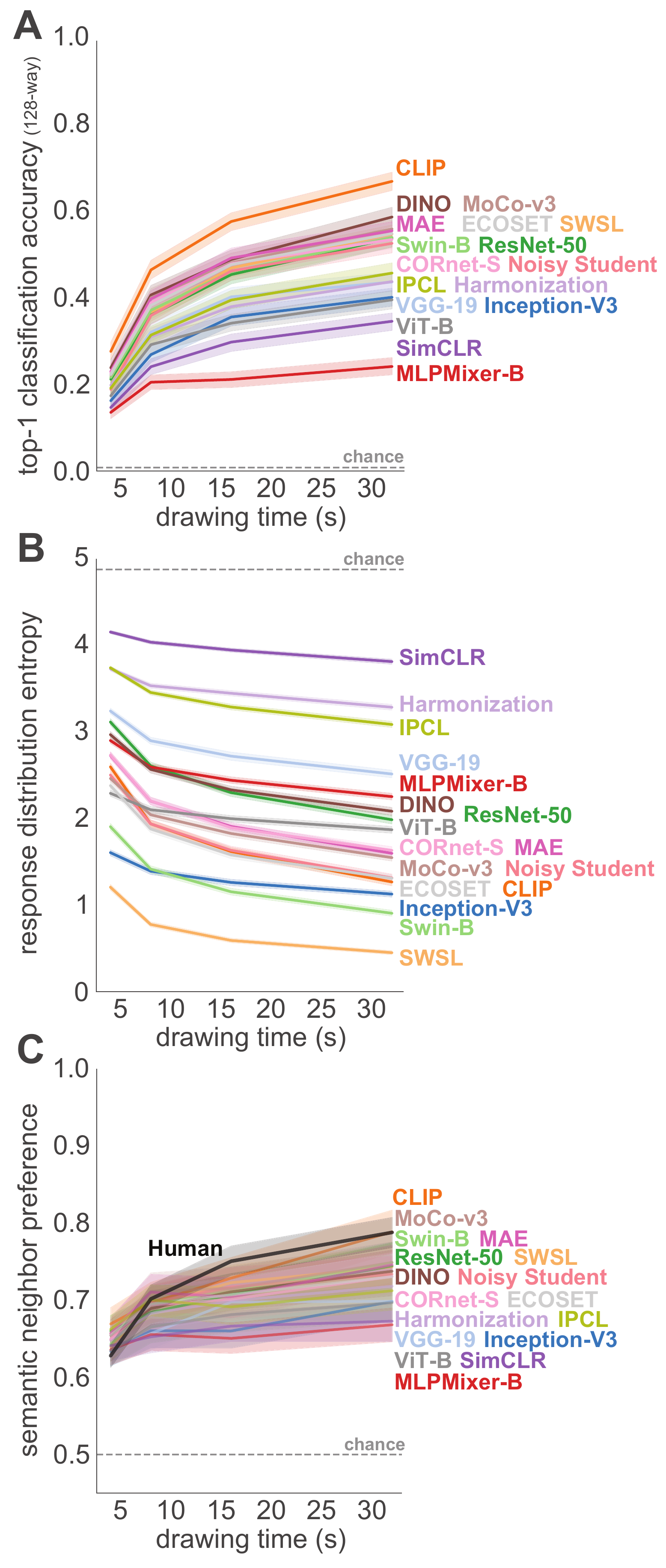}
    \caption{Effect of drawing time constraints on sketch understanding in different vision models. }
    \label{fig:accuracy_entropy_snp}
    \vspace{-1em}
\end{wrapfigure}

\textbf{Sparser sketches are more semantically ambiguous for models and humans.}
Having verified that we had successfully manipulated the level of detail in human sketches, we next sought to evaluate how well current vision models extract semantic information from them at each level of detail. 
Our general approach was to fit mixed-effect linear regression models to estimate the effect of sparsity on 3 key metrics: (1) top-1 classification accuracy, (2) entropy of response distributions, and (3) semantic neighbor preference. Regression models included random intercepts and slopes for object concepts.
Figure \ref{fig:accuracy_entropy_snp} shows the performance of each vision model with respect to these metrics for sketches produced under different time constraints.

% We also estimated whether different vision models had comparable performance with respect to each of these metrics. 
We found that models generally achieved higher top-1 classification accuracy for more detailed sketches than sparser ones ($\beta=9.57\times10^{-2}$, $t=20.25$, $p<.001$).
% (A) top-1 classification accuracy (B) response distribution entropy and (C) semantic neighbor preference

We further found the entropy of the models' response distribution was lower for detailed sketches than for sparser sketches ($\beta = -.03$, $t = -30.19$, $p<.001$), suggesting greater uncertainty about the best label to apply to sparser sketches. 
Even when sketches were more ambiguous, however, models generated labels that were semantically related to the ground-truth label, as measured by our \textit{semantic neighbor preference} score, with more detailed sketches eliciting a greater proportion of semantically related labels ($\beta = 2.5 \times10^{-2}$, $t=8.43$, $p <.001$).
These patterns were mirrored in human sketch recognition behavior, with more detailed sketches being associated with higher top-1 classification performance ($\beta = 6.08\times10^{-2}$, $t=9.57$, $p<.001$), a tighter distribution of responses (lower entropy) ($\beta=-.14$, $t=-12.29$, $p<.001$), and greater semantic neighbor preference ($\beta=5.41\times10^{-2}$, $t=20.24$, $p<.001$).

% We characterized sketch understanding in humans and all models using a combination of three metrics: (1) top-1 classification accuracy, reflecting raw sketch recognition performance; (2) Shannon entropy of the response distribution, reflecting the degree of uncertainty about the target label; and (3) \textit{semantic neighbor preference (SNP)}, reflecting the degree to which models and humans generated off-target responses that were semantically related to the target label. 
% Are vision models' abilities to understand the semantic content depicted in sketches influenced by the sparsity of the sketches?
% We characterized sketch understanding using a combination of three metrics: (1) top-1 classification performance, (2) the entropy of a model's response distribution over labels for each sketch, and (3) \textit{semantic neighbor preference}, the likelihood of a model to provide labels that are semantic neighbors of the ground-truth label even when their top-1 guess is incorrect.
% We also measured human performance across these 3 metrics to see the effect of sketch sparsity on human sketch understanding.

% \textit{Top-1 classification accuracy.} 

\textbf{Different models display distinct patterns of sketch recognition behavior.}
Although all vision models were sensitive to the effect of our drawing-time manipulation, we found that there were reliable differences in classification accuracy between models ($\chi^2(16) = 3455.3$, $p<.001$). % , as revealed by nested statistical model comparison 
Moreover, some models generated a greater diversity of responses than others, as measured by the entropy of their response distribution (Fig.~\ref{fig:accuracy_entropy_snp}B, $\chi^2(16) = 89698$, $p<.001$). %, indicating that some models generated a greater diversity of responses than others.
Finally, models varied in the degree to which they generated non-ground-truth labels that were semantically related to the ground-truth label ($\chi^2(16)=318.46$, $p<.001$). 
Taken together, these results indicate that these models, all high-performing, display systematic differences in how they extract semantic information from sketches.

\begin{figure}[htp!]
    \centering
    \includegraphics[width=.75\textwidth]{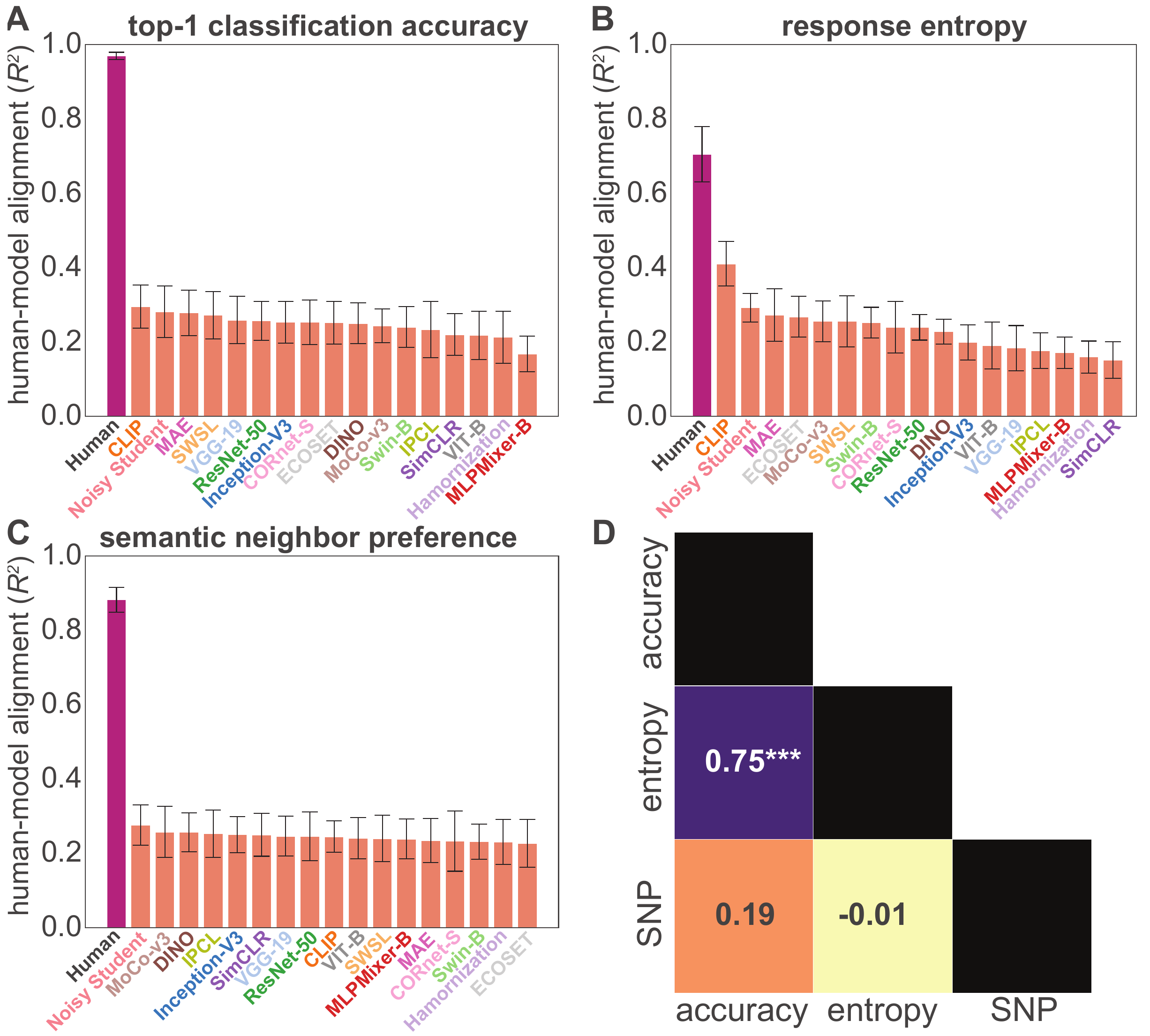}
    \caption{Human-model alignment on (A) top-1 classification accuracy, (B) response entropy, and (C) semantic neighbour preference. Leftmost red bars in each plot correspond to baseline human-human consistency on each metric. Error bars indicate bootstrapped 95\% confidence intervals. (D) Spearman $\rho$ correlations between the rank-ordering of vision models with respect to their alignment to human performance on each metric.}
    \label{fig:alignment}
    \vspace{-1em}
\end{figure}
\textbf{A large gap remains between human and model sketch understanding.}
While both humans and models are affected by the amount of detail in sketches, it is not yet clear to what degree their response patterns are well aligned. 
% show similar variations in sketch understanding across levels of detail, are they aligned along these metrics at a more granular level of analysis?
We evaluated human-model alignment scores using the same three metrics (i.e., top-1 classification accuracy, entropy, semantic neighbor preference) by estimating the degree to which model performance on different \textit{types} of sketches (e.g., lions drawn in 4 seconds or less) covaried systematically with human performance on the same \textit{types} of sketches. 
For example, a model is considered well aligned with humans with respect to recognition performance if it achieves high top-1 classification accuracy on the same types of sketches that humans succeed in classifying \textit{and} if it achieves low accuracy on the types of sketches that humans fail to classify. 
Similarly, a model is considered well aligned with humans with respect to semantic ambiguity if it produces a response distribution with high entropy for the same types of sketches that humans are also highly uncertain about \textit{and} if it produces a low-entropy response distribution for the types of sketches that humans systematically agree on (regardless of whether this agreement is concentrated on the correct label).
% \end{wraptable}
% We sought to test how aligned humans and models were on top-1 classification accuracy, response entropy, and semantic neighbor preference for each \textit{kind} of sketch. 
% Here, \textit{kind} refers to sketches of a given concept category at a given level of detail. 
% Our dataset had 128 concepts $\times$ 4 draw duration conditions, furnishing 512 kinds of sketches.
% We fit linear mixed-effects models predicting model performance from human performance along our 3 metrics. We included a fixed-effect interaction term for drawing duration to account for variance across the different levels of sparsity and a fixed-effect term for vision-model type to account for variations in human-model alignment across vision models.
We found that models generaly displayed some degree of alignment to humans on top-1 classification accuracy ($\beta= 7.70\times10^{-2}$, $t=6.77$, $p<.001$), response entropy ($\beta=1.17\times10^{-1}$, $t=28.52$, $p<.001$), and semantic neighbor preference ($\beta = 6.66\times10^{-2}$, $t = 6.21$, $p<.001$). 
Moreover, we found that different vision models aligned with humans to varying degrees (top-1 classification accuracy: $\chi^2(16) = 134.81$, $p<.001$; response entropy: $\chi^2(16) = 725.78$, $p<.001$; semantic neighbor preference: $\chi^2(16) = 5.05$, $p=.99$).
% $\chi^2$ model comparisons between models with and without a fixed-effect term for vision model-type revealed that while there was significant variation across vision models with respect to human-model alignment for top-1 classification accuracy ($\chi^2(16) = 134.81$, $p<.001$) and response entropy ($\chi^2(16) = 725.78$, $p<.001$), all models showed similar alignment for semantic neighbor preference ($\chi^2(16) = 5.05$, $p=.99$).
Nevertheless, a sizable gap remains between the most aligned models and a human-human consistency baseline for \textit{all} metrics (Fig~\ref{fig:alignment}; top-1 classification accuracy: $t = 952.19$, $p<.001$; response entropy: $t = 184.21$, $p<.001$; semantic neighbor preference: $t=389.56$, $p<.001$).
Finally, we observe that while examining classification accuracy and entropy yield similar rankings over which models are best aligned to humans, these metrics appear to capture non-redundant sources of information about human and model sketch understanding (Fig~\ref{fig:alignment}D).

\textbf{A CLIP-based sketch generation algorithm emulates human sketches under some conditions.}

\begin{wrapfigure}{r}{.6\textwidth}
    \centering
    \includegraphics[width=.6\textwidth]{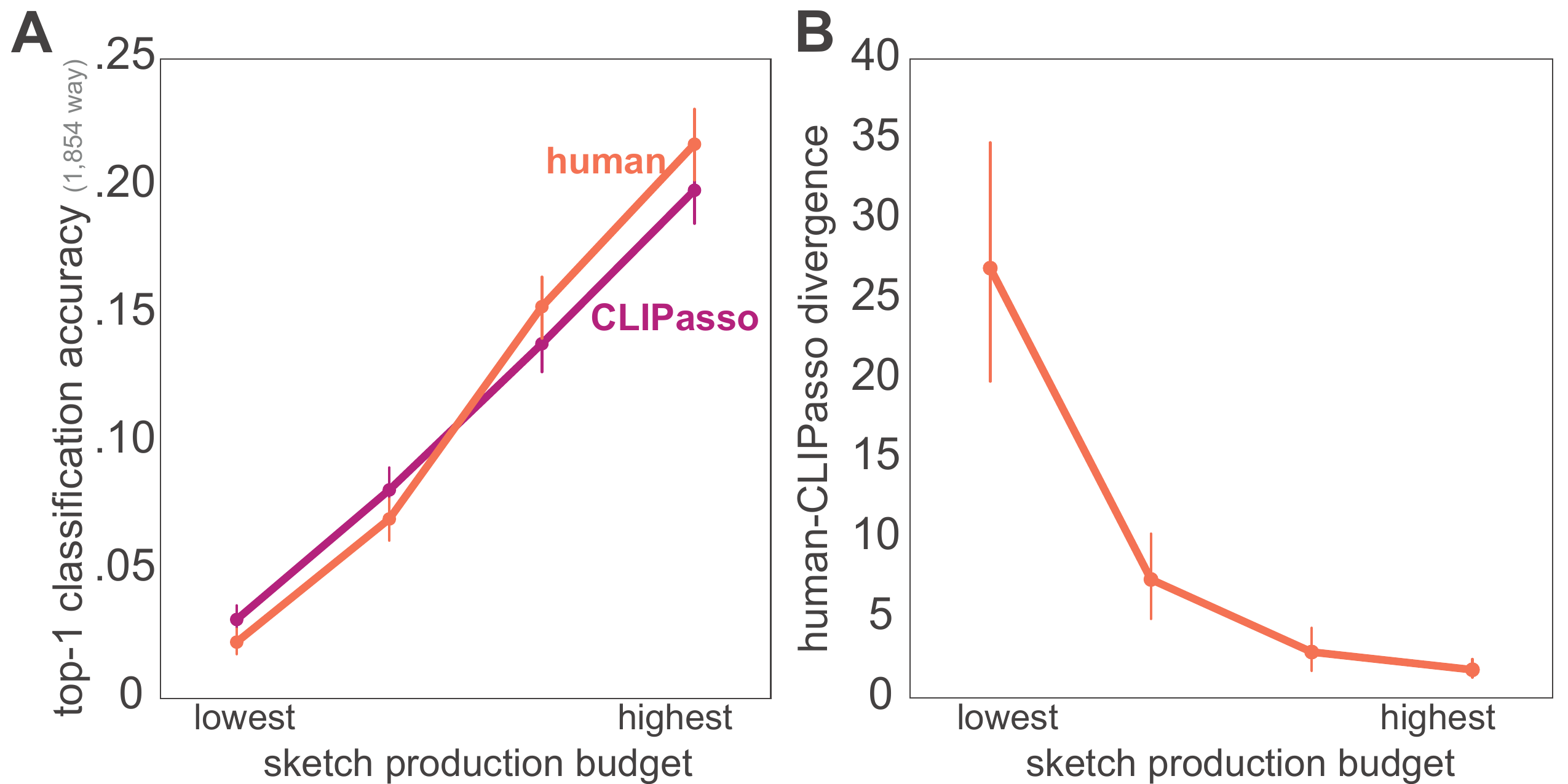}
    \caption{(A) Human top-1 recognition performance on \texttt{CLIPasso} and human sketches at different production budgets (draw duration for human sketches and number of strokes for \texttt{CLIPasso} sketches). (B) Divergence in human-responses to the same type of human and \texttt{CLIPasso} sketches as a function of sketch production budget.}
    % \vspace{-1em}
    \label{fig:CLIPasso-results}
\end{wrapfigure}

While sketch understanding is a critical aspect of visual abstraction, the ability to \textit{produce} sketches spanning different levels of abstraction is no less important.
Although there remains a gap between human and model sketch understanding, a CLIP-based vision model \cite{radford2021learning} was among the most performant and best aligned to human sketch understanding. 
% our results thus far highlight the promise of CLIP, one of the models that displayed the strongest alignment with human sketch understanding. 
% a large vision-language model trained on web-scale data, CLIP, is among the most performant models at sketch recognition without any fine-tuning and also one of the most well aligned models to human sketch understanding.
Insofar as the major bottleneck to being able to generate more human-like sketches is achieving more human-like understanding of sketches and other images \cite{fan2018common}, a generative model leveraging CLIP's latent representation may be a promising approach. % could a generative model that leverages CLIP's latent features generate sketches at multiple levels of visual abstraction?
Consistent with this possibility, we found that \texttt{CLIPasso} generated sketches of concepts at each abstraction level were about as recognizable as the human-generated sketches of those same concepts at the same abstraction level (Fig.~\ref{fig:CLIPasso-results}A; $adjusted \: R^2 = .64$).
% human sketches of a given concept at a particular abstraction level (e.g., lions drawn in 32s) that were 
% As can be seen in Figure (XXX) human recognition accuracy for human sketches and CLIPasso sketches were comparable at each level of `production budget' (draw duration for human sketches, number of strokes for CLIPasso sketches).
% We fit a mixed-effect linear regression model predicting top-1 classification accuracy of human sketches from top-1 classification accuracy of the same \textit{kind} of CLIPasso sketches with an interaction term for abstraction level (number of strokes).
% Overall, we found a close correspondence between people's recognition performance on human and CLIPasso performance ($adjusted \: R^2 = .64$).
Moreover, we found that the more detailed \texttt{CLIPasso} sketches were especially human-like in that they evoked a similar set of meanings to their human-generated counterparts.
% Since human recognition performance on CLIPasso sketches does not perfectly account for performance on human drawings, this indicates that while humans see \textit{some} similar structure between human-made and CLIPasso-generated sketches of the same concept at each level of abstraction, there are some sketches that lead to diverging human responses depending on whether they are made by humans or CLIPasso. 

To measure the degree to which human and \texttt{CLIPasso} sketches converged with respect to the object labels that they elicited from human viewers, we computed the Jensen-Shannon distance (JSD) between the label distributions of human and \texttt{CLIPasso} sketches for each concept at each of the 4 levels of abstraction.
% We compared the vector of human label responses between CLIPasso and human sketches of each concept at each level of abstraction using the Jensen-Shannon distance (JSD) between the normalized human-generated label counts. 
In Fig. \ref{fig:CLIPasso-results} B. we show the average label divergence at each level of abstraction.
% For each \textit{kind} of CLIPasso sketch, kind referring to sketches of each concept at each of the 4 constraint levels, we ranked each kind of human sketch in terms of their JSDs.
% We computed the minimum rank amongst human sketches of the same concept at each level of constraint for CLIPasso sketches (number of strokes).
% This captures the minimum divergence between the target CLIPasso sketch and \textit{any} human sketch of the same concept regardless of level of abstraction.
% We found that more detailed sketches, i.e., those with a greater number of strokes, tended to have an overall higher rank across concepts.
We found that humans and \texttt{CLIPasso} sketches were least divergent in terms of their perceived meaning when they were depicted in greater detail or were \textit{less abstract} ($\beta = -0.69$, $t=-6.86$, $p<.001$).
At lower levels of detail and visual fidelity, human and \texttt{CLIPasso} sketches elicited more diverging responses.
Thus, while \texttt{CLIPasso} more closely approximates human sketch production behavior at greater levels of detail, there remains a large gap between how \texttt{CLIPasso} and human participants attempted to preserve sketch meaning under more severe production constraints. 
Taken together, while \texttt{CLIPasso} marks significant progress towards human-like sketch production its ability to produce highly abstract sketches in a human-like manner remains limited.

\newpage
\section{Conclusion}
\vspace{-1 em}
Recent advances in machine vision have enabled new opportunities to understand the computational mechanisms that underlie human-like visual abstraction---how humans create and interpret a wide variety of images (from detailed illustrations to schematic diagrams) to convey what they perceive and know about the world. 
% There has been incredible progress in recent years in the development of high performing vision models, which have steadily achieved enough variation in model architecture and training to prime cognitive psychology research with the opportunity to begin explore the extent to which specific models have achieved human-like understanding of abstract symbolic representations, like sketches.
Here we introduce \texttt{SEVA}, a new dataset and benchmark containing approximately 90K human and machine generated sketches spanning multiple levels of abstraction, to evaluate progress towards alignment between human and machine visual abstraction. 
 % Critically, our dataset is the first sketch dataset to include systematic variation in sparsity in order to test state-of-the-art vision models in their ability to achieve human-like understanding of visual concepts across multiple levels of abstraction. 
Critically, our model evaluation protocol is focused not just on performance but on how well these models approximate \textit{human-like} sketch production and understanding.
Initial benchmarking of a set of 17 vision models on \texttt{SEVA} following this protocol revealed that even the most performant and well-aligned models deviate from human behavior in systematic ways.
We also find that current generative models of sketching are able to sketch in human-like ways, but only in limited settings. 

We hope that \texttt{SEVA} will accelerate progress towards unified computational theories that explain how humans are capable of generating and understanding such a wide variety of abstract visual representations.
Specifically, our dataset could be used to adjudicate between vision models that are high-performing on tasks involving natural images \cite{mahmoud2023stress,golan2020controversial} by characterizing their alignment with how humans understand images generated in a very different manner, including freehand sketches. 
In addition, \texttt{SEVA} is distinctive among existing human sketch datasets in that it contains detailed measurements of how humans make moment-to-moment decisions about where to place each stroke, when aiming to produce a sketch of a visible object, under different constraints. 
As such, we expect \texttt{SEVA} to be a key resource for advancing the state-of-the-art in sketch generation \cite{qiu2022emergent,wang2021sketchembednet, vinker2022clipasso,ha2017neural,ribeiro2020sketchformer}, and thus lead to computational models of generalized visual abstraction. 

\vspace{-1em}
\section{Limitations}
\vspace{-1em}
We note several limitations of the current study that would be important to address in future work. 
\underline{First}, our sample of participants was limited to English-speaking individuals based in the United States. As such, the current study cannot speak to potential differences in sketch-production behavior across geographical and cultural contexts. 
However, future work that recruits from a broader cross-section of individuals, including those located in the ``Majority World,'' will be vital for understanding those potential sources of variation in human sketch production and comprehension. 
\underline{Second}, we recruited participants via Prolific, a widely used crowdsourcing platform in human behavioral research, without regard to any previous artistic training they had received. As such, our study generally reflects sketch production behavior among individuals without substantial expertise in the visual arts.
\underline{Third}, following prior work \cite{fan2018common, fan2020pragmatic, hawkins2023visual,yu2016sketch,jongejan2017quick, long2021parallel}, the human sketches in our dataset were obtained using a web-based digital drawing interface, where most participants used a mouse or trackpad to produce their sketches (89.96\%) and some participants used a touchscreen (6.94\%) or a stylus (1.67\%). 
Investigation of the impact of input device on sketch production would be a fruitful avenue for follow-up work. 
\underline{Fourth}, because we did not obtain non-digital sketches (e.g., drawn on paper with a pen/pencil), our data cannot speak to differences between digital and non-digital sketches.
\underline{Fifth}, our study of machine sketch production included just one model, \textit{CLIPasso}, limiting the conclusions that can be drawn about sketch generation algorithms in general. Future work that evaluates a broader suite of algorithms would thus be valuable. 

\newpage
\section*{Acknowledgements}
We thank members of the Cognitive Tools Lab at Stanford and the Stanford Neuro AI Lab for helpful feedback, discussions, and support.
This work was supported by an NSF CAREER Award \#2047191 to J.E.F..
J.E.F is additionally supported by an ONR Science of Autonomy award.
\vspace{1em}

\vspace{1em}
\fbox{\parbox[b][][c]{\textwidth}{\centering {All code and materials available at: \\
\href{https://github.com/cogtoolslab/visual_abstractions_benchmarking_public2023/}{\url{https://github.com/cogtoolslab/visual_abstractions_benchmarking_public2023/}}
}}}

\section*{Broader Impacts}

Visual abstraction is as ubiquitous as it is central to our understanding of the visual world.
Humans can recognize depictions of objects at varying levels of fidelity to their real-world counterparts. 
This ability to recognize and render concepts in abstract forms is crucial for visual knowledge transmission in the form of graphs, diagrams, and symbols.
While computer vision has steadily made progress towards algorithms that can recognize objects in scenes, distinguish among different instances of an object, or even answer questions about those objects in natural language, it remains unclear if these systems have the capability to understand visual concepts at multiple levels of abstraction in a human-like manner.
\texttt{SEVA} marks a step towards evaluating current vision algorithms on their sensitivity to depictions of a wide variety of common object concepts at varying levels of abstraction.
Critically, we also provide results on \textit{human} performance on sketch understanding for sketches of varying sparsity, setting a clear benchmark for future vision algorithms.
Our initial study of state-of-the-art vision algorithms shows that even the most high-performing of models still fall short of human consistency baselines.
With these vision algorithms increasingly being deployed in many aspects of everyday life, it is crucial to measure how human-like they are in this most natural of human abilities. 
We hope that \texttt{SEVA} will be used by the community to help close the gaps in current vision algorithms' alignment to human behavior that we have identified in this paper.
% Visual abstraction is a powerful tool that enables humans to distill complex information, ranging from diagrams and flowcharts to the diverse symbols found in city highways, into a simplified and accessible form.
% Despite extensive progress on algorithmic understanding and creation of images and scenes, the capacity of these models to effectively grasp varying levels of visual abstraction remains an open question in many AI applications, including the creation of graphics and logos, data visualization and infographics, user interface design, visual storytelling and animation, and other fields that require condensing information into visual forms. As such, we believe the SEVA benchmark is a key step toward measuring whether a given algorithm is sensitive to visual abstraction akin to human capabilities. In the case that sensitivity to visual abstraction becomes a crucial aspect of achieving high performance in some real-world domain, SEVA (or its successors) could be used to screen for algorithms more likely to generalize to a wide range of visual representations. Furthermore, our results, while representative of only an initial survey of existing algorithms, do suggest that models with a symbolic understanding of the world (e.g. CLIP-ViT) better mirror human comprehension of visual abstraction, suggesting that symbolic representations may be key to human-like inferences about visual abstraction for vision algorithms.

\bibliography{references}{}
\bibliographystyle{plain}

\newpage
\appendix

\end{document}

\section{Supplementary Materials}
We confirm that the dataset will be available to the community under an MIT licence and that we bear all responsibility for the dataset and the associated codebase including in the case of violation of rights.
\subsection{Code and Materials}

Code for reproducing the analyses reported in this submission as well as links to the dataset can be found here. \\{\url{https://github.com/cogtoolslab/visual_abstractions_benchmarking_public2023}}

% Materials including the dataset can be found here - \\{\url{https://www.dropbox.com/scl/fo/2oqncsagow0k7sbn52pd1/h?dl=0&rlkey=i7ezf9lezft7o0amawb24zlvd}}

\subsection{Datasheet for Dataset}
\subsubsection{Motivation}
\begin{itemize}
    \item \textbf{For what purpose was the dataset created?}
    To measure both humans and vision models in their ability to understand visual concepts at varying levels of abstraction.
    \item \textbf{Who created the dataset and on behalf of which entity?} The dataset was composed by the authors of this paper, and the data was generated by workers on Prolific and a generative sketching model, CLIPasso.
    \item \textbf{Who funded the creation of the dataset?} Funding for data collection was provided by NSF CAREER #2047191 to J.E.F. 
    
\end{itemize}
\subsubsection{Composition}
\begin{itemize}
    \item \textbf{ What do the instances that comprise the dataset represent?} Each instance consists of a digital sketch of one of 128 objects drawn under different time constraints. Some sketches were generated by a generative sketching model under different stroke constraints.
    \item \textbf{How many instances are there in total?} There are 89,797 individual human sketches. There are also 8,192 machine-generated sketches.
    \item \textbf{Does the dataset contain all possible instances or is it a sample of instances from a larger set?} Given the space of possible drawings at different abstraction levels is infinitely large, our dataset does not encompass every possible sketch instance. However, we provide many sketches of each concept at each abstraction level so as for the dataset to be representatively diverse.
    \item \textbf{What data does each instance consist of?} The sketch data consists of a *.png image of each sketch along with information about stroke trajectory, undo-history, the time-constraint or stroke-constraint under which it was produced, a label for the object it represents and which of 16 photographs of that object was shown to a participant when they were asked to draw it.
    Some sketches also have associated recognition data. This includes labels provided by human participants regarding what object they thought the sketch represented.
    \item \textbf{ Is there a label or target associated with each instance?} Yes, each sketch is associated with a concept label and an abstraction level label.
    \item  \textbf{Is any information missing from individual instances?} No, not that we are aware of. If we find any missing information we will update the dataset accordingly.
    \item \textbf{ Are relationships between individual instances made explicit?} Yes, all relevant metadata to make connections between instances are provided.
    \item \textbf{Are there recommended data splits?} No.

    \item \textbf{Are there any errors, sources of noise, or redundancies in the dataset?} It is likely that some sketches might not be informative or might be noise. As we discover such instances, we will update our dataset accordingly.

    \item \textbf{Is the data self-contained, or does it link to or otherwise rely on external resources? } It is self-contained.

    \item \textbf{Does the dataset contain data that might be considered confidential?} No.
    
    \item \textbf{Does the dataset contain data that, if viewed directly, might be offensive, insulting, threatening, or might otherwise cause anxiety?} No.
    \item \textbf{Does the dataset relate to people?} Most of the sketches were generated by human participants, but otherwise no.

\end{itemize}
\subsubsection{Collection Process}
\begin{itemize}
    \item \textbf{How was the data associated with each instance acquired? What mechanisms or procedures were used to collect the data? How was it verified?} Sketches were produced by human workers on Prolific who were compensated for their time. The data were collected online using a JavaScript based sketching tool built in jsPsych.
    Recognition data were also collected on Prolific using another JavaScript-based tool also built in jsPsych. 
    After accepting the task, participants provided informed consent in accordance with our university IRB and were provided instructions on how to do the task.
    Data were verified by the experimenters after data collection was completed
    \item \textbf{Who was involved in the data collection process and how were they compensated?} The data were collected by the authors of the paper and data were generated by Prolific workers. They were compensated at the rate of \$15.50/hour according to the California minimum-wage.
    \item \textbf{Over what timeframe was the data collected?} All data were collected between January and May 2023 at different times.
    \item \textbf{Were any ethical review processes conducted?} All human data collection was approved by the University of California, San Diego IRB.
    
\end{itemize}
\subsubsection{Preprocessing/cleaning/labeling}
\begin{itemize}
    \item \textbf{Was any preprocessing/cleaning/labeling of the data done (e.g.,
discretization or bucketing, tokenization, part-of-speech tagging,
SIFT feature extraction, removal of instances, processing of missing values)? }
The raw data we present in our dataset were not put through any preprocessing pipelines. We did exclude data from participants who faced technical difficulties in our online task.
A subset of our data were shown to a new cohort of Prolific workers for a recognition experiment and we provide data associated with that experiment in our dataset.
\end{itemize}
\subsubsection{Uses}
\begin{itemize}
    \item \textbf{Has the dataset been used for any tasks already? } Yes, the sketches produced in the human and machine sketch generation experiments were shown to participants to collect recognition data. In this paper, we also conduct several analyses on these data.
    \item \textbf{Is there a repository that links to any or all papers or systems that use the dataset?} No other papers use the dataset yet.
    \item \textbf{What (other) tasks could the dataset be used for?} The dataset could be used to finetune or adapt existing vision models and could also be used for human behavioral experiments that require participants to recognize abstract sketches.
    \item \textbf{Is there anything about the composition of the dataset or the way it was collected and preprocessed/cleaned/labeled that might impact future uses?} Not that we are aware of.
    \item \textbf{Are there tasks for which the dataset should not be used? } The authors are unaware of any use cases of this dataset that should be prohibited.
\end{itemize}
\subsubsection{Distribution}
\begin{itemize}
    \item \textbf{Will the dataset be distributed to third parties outside of the entity (e.g., company, institution, organization) on behalf of which the dataset was created? } Yes.
    \item \textbf{How will the dataset will be distributed?} It will be available using a combination of public Github repositories and links to raw human data on Amazon S3 instances.
    \item \textbf{When will the dataset be distributed?} During the time of the submission of this manuscript.
    \item \textbf{Will the dataset be distributed under a copyright or other intellectual property (IP) license, and/or under applicable terms of use (ToU)? } The dataset and associated code will be licensed under the MIT license.
    \textbf{Have any third parties imposed IP-based or other restrictions on the data associated with the instances?} No.
    \textbf{Do any export controls or other regulatory restrictions apply to the dataset or to indi- vidual instances?} No.
\end{itemize}
\subsubsection{Maintenance}
\begin{itemize}
    \item \textbf{Who is supporting/hosting/maintaining the dataset?} The dataset will be maintained by researchers at the Cognitive Tools Lab at Stanford University.
    \item \textbf{How can the owner/curator/manager of the dataset be contacted?} The corresponding author for this paper can be reached at their email. Links to the email will also be available on the public GitHub repository.
    \item \textbf{Is there an erratum?} Not presently.
    \item \textbf{ Will the dataset be updated (e.g., to correct labeling errors, add new instances, delete instances)?} Yes, the dataset will be updated to correct for any errors.
    \item \textbf{Will older versions of the dataset continue to be supported/hosted/maintained?} Yes, as long as there aren't harmful or offensive sketches in the older versions.
    \item \textbf{If others want to extend/augment/build on/contribute to the dataset, is there a mechanism for them to do so? } We will provide code for our online sketching task and recognition task. If there is interest in building upon the dataset, the community will have the tools to do so. The authors welcome correspondence on this topic.

\end{itemize}

\subsection{Human Sketch Production Task}

\begin{figure}[htp]
    \centering
    \includegraphics[width=.75\textwidth]{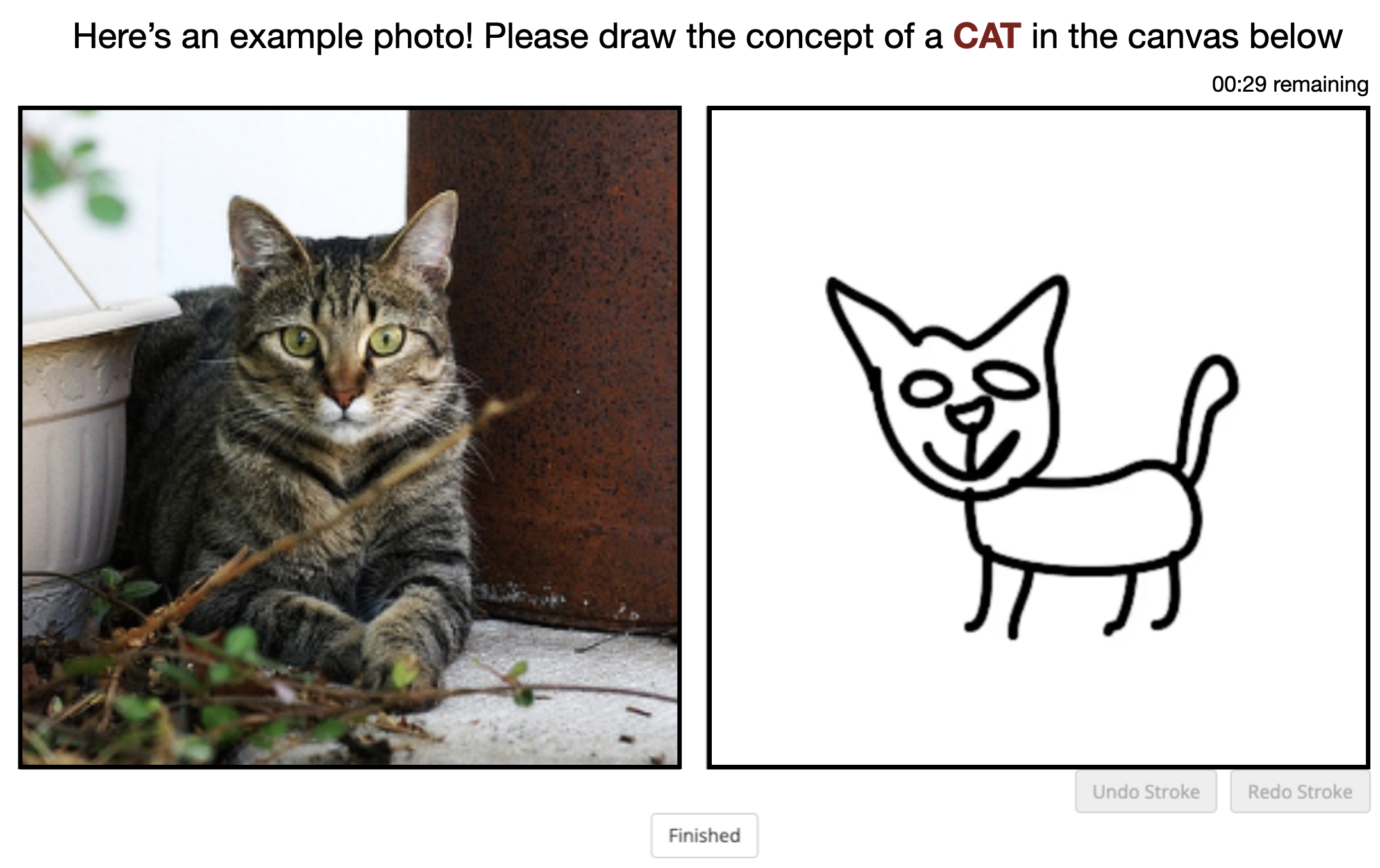}
    \caption{Web display for human sketch production task.}
    \label{fig:production_task}
\end{figure}

In each trial, participants were presented with a color photograph (500px x 500px) of an object paired with its concept label and asked to produce a drawing of the object concept in either 4 seconds, 8 seconds, 16 seconds, or 32 seconds. See Fig. \ref{fig:production_task} for an example task display.

Prior to drawing, participants were told that the photograph cue was provided as a reminder of what the object looked like, but that they should make their drawings as recognizable as they could from the concept level, rather than of that specific example.
They were encouraged to use as much time as they were permitted to produce as recognizable of a drawing as they could. 
The trial ended either when participants submitted their drawing or when time ran out. 

\begin{figure}[ht!]
    \centering
    \includegraphics[width=.85\textwidth]{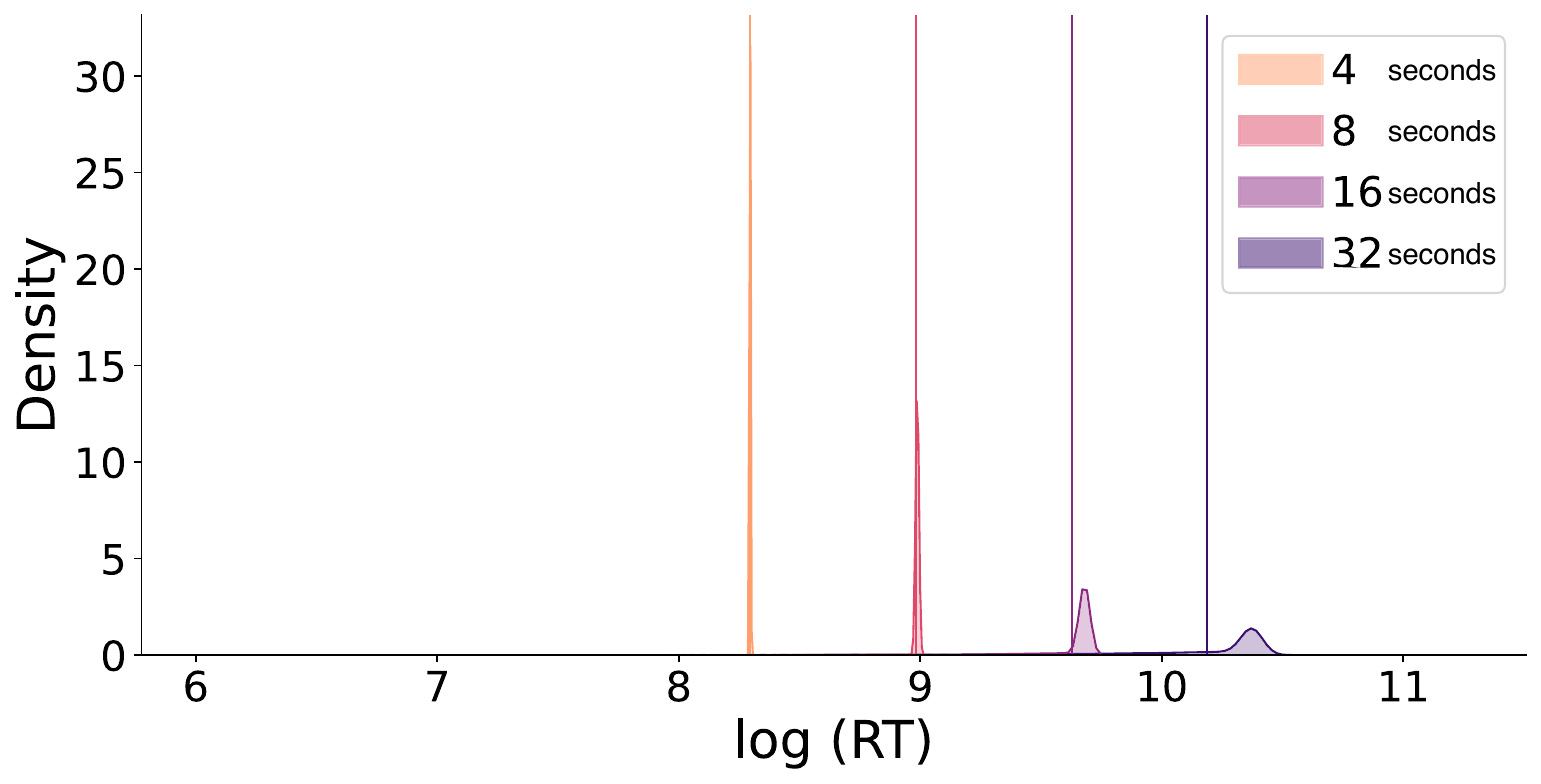}
    \caption{Distrubtion of Log response times (RTs) for each drawing constraint condition. Lines indicate mean log RTs. }
    \label{fig:rt_by_abstraction}
\end{figure}
\begin{figure}[ht!]
    \centering
    \includegraphics[width=.85\textwidth]{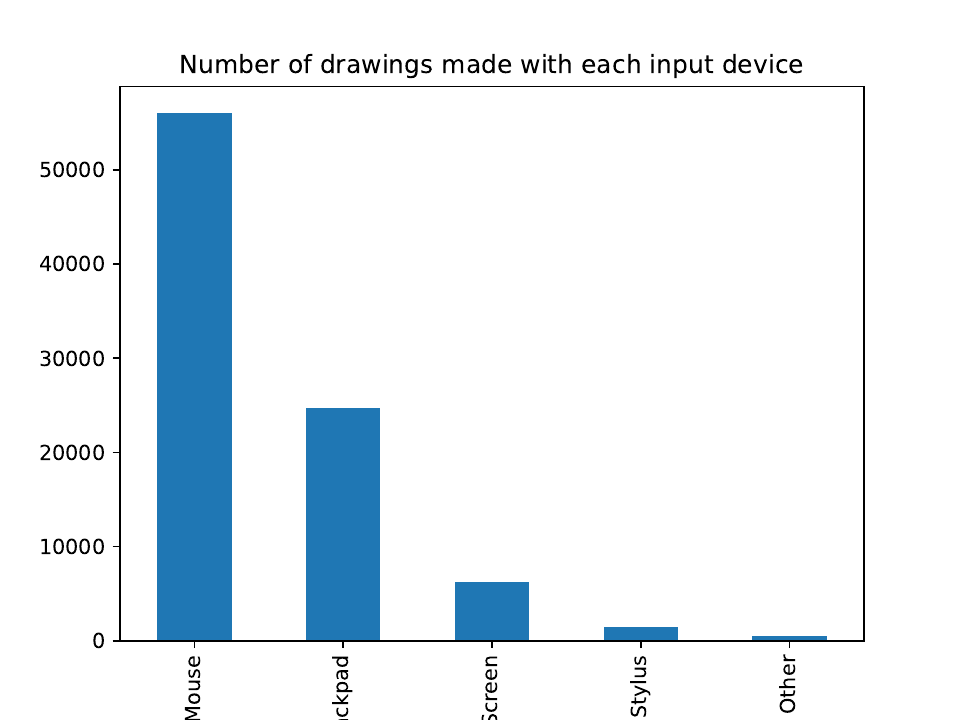}
    \caption{Number of sketches made using each unique input device. }
    \label{fig:devices}
\end{figure}

\begin{figure}[ht!!]
    \centering
    \includegraphics[width=.85\textwidth]{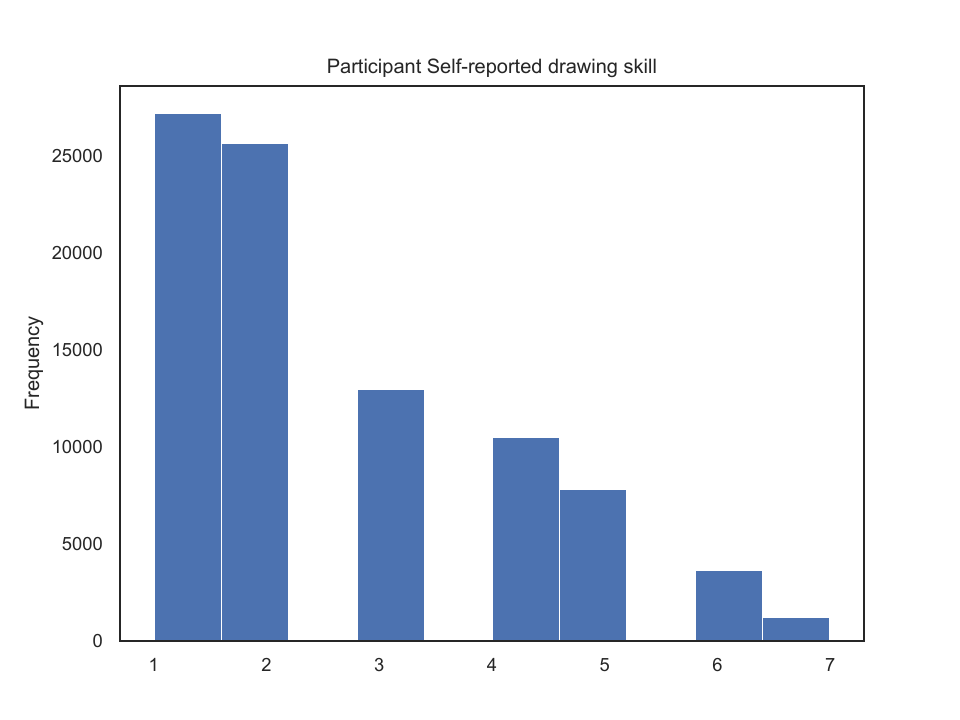}
    \caption{Distribution of participant reported drawing skill (on a scale from 1 to 7). }
    \label{fig:devices}
\end{figure}

\subsection{Human Sketch Recognition Task}
\vspace{-1em}
\begin{figure}[ht!]
    \centering
    \includegraphics[width=.75\textwidth]{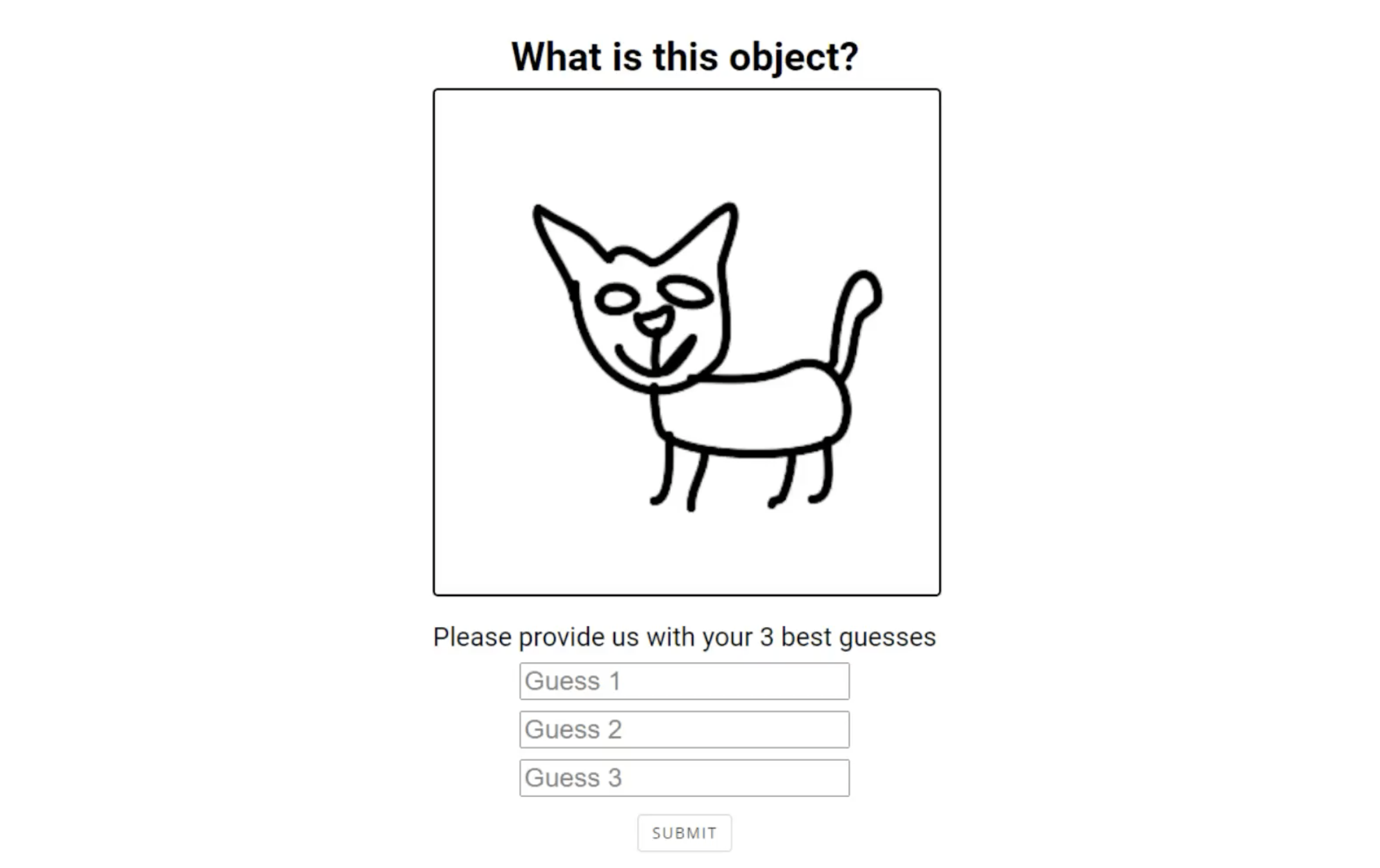}
    \caption{Web display for human sketch recognition task.}
    \label{fig:recognition_task}
\end{figure}

On each trial of the recognition task, participants were presented with a single sketch on a canvas along with 3 blank text boxes.\ref{fig:recognition_task}. They were asked "What is this object?" and asked to provide at least 1 guess.
Upon beginning to type their guess in any of the text boxes, a set of labels would appear in dropdown menu that contained matches from the entire set of 1,854 labels in the \texttt{THINGS} dataset to the current set of characters typed.
Only responses that matched one of these 1,854 labels were accepted as valid.

\subsection{Model Evaluation Protocol}
\subsubsection{Extracting Class Probabilities from Vision Models}
In this section, we provide a detailed description of the model evaluation protocol used in our study. For each pre-trained vision encoder in our model suite, we fit a regularized logistic regression model to predict the class label of each sketch based on the latent features of the model. 

Following common practice in linear evaluation, we excluded the linear classification head from each pre-trained model and generated latent features using the rest of the encoder. In cases where the extracted features contained spatial dimensions (e.g., $height \times width \times channels$, or $tokens \times channels$), we computed the mean along the spatial dimensions, resulting in a 1-D vector feature for each sketch.

Then we used 5-fold stratified cross validation to assess the performance of the models. The human drawing dataset was divided into five splits, with the same number of $class \times production\_condition$ in each split. We fit a logistic regression model on each training split and used it to predict the labels of sketches in the corresponding test split. We used the logistic regression implementation from scikit-learn, with L2 regularization and L-BFGS optimizer. 

Due to resource limitations, we performed a parameter search for the regularization strength on CLIP and VGG-19 models, over \{0.01, 0.1, 1.0, 10.0, 100.0\} Both models achieved the best performance with a regularization strength of 1.0. Therefore, we applied this parameter to the entire suite of models. We acknowledge that utilizing a fixed parameter may not yield the optimal performance for all models. In future versions of this paper, we plan to conduct individual parameter searches for each model to ensure the best possible performance across the board.

\subsubsection{Details On Semantic Neighbor Preference Score }
\begin{figure}
    \centering
    \includegraphics[width=.9\textwidth]{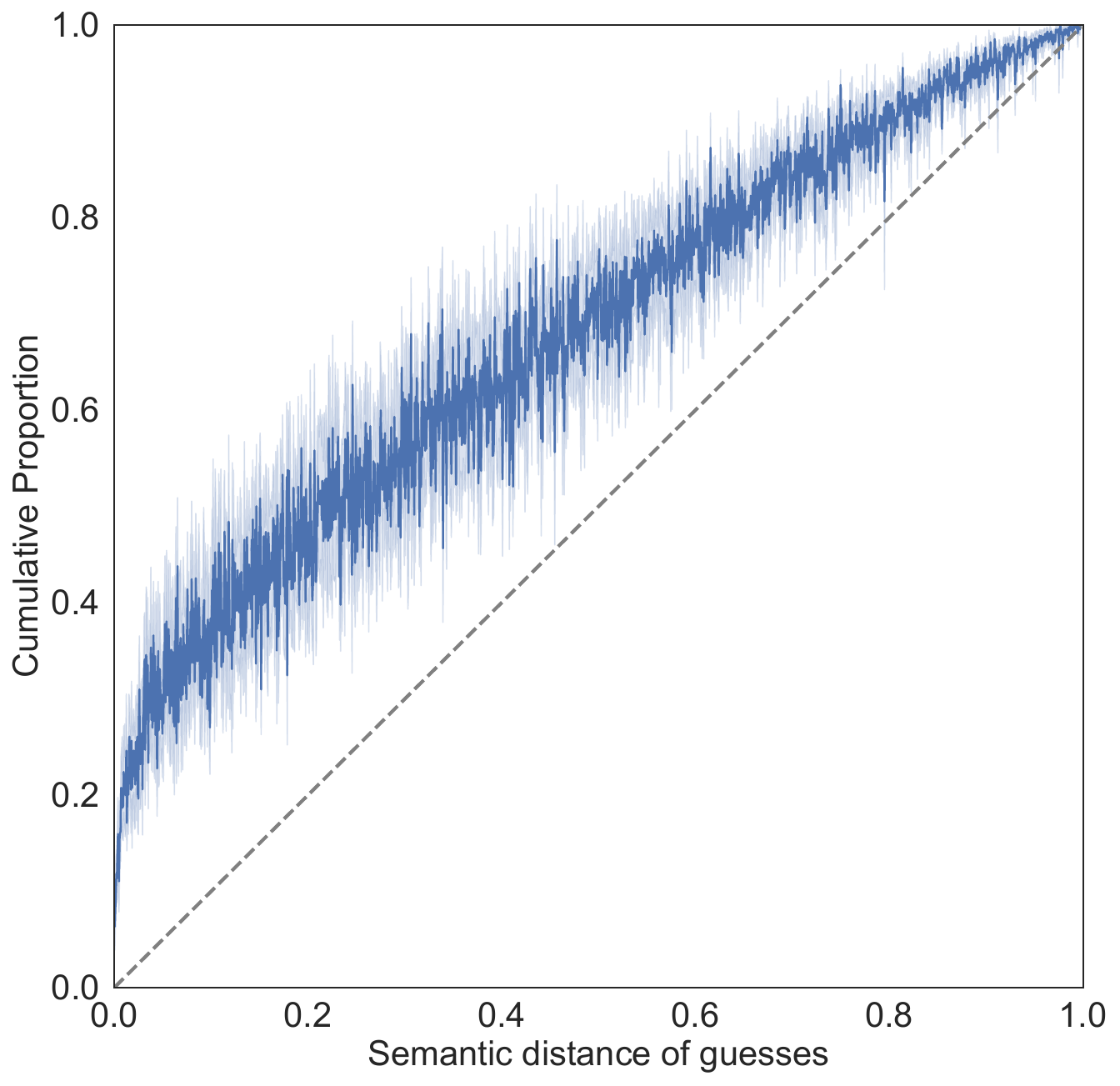}
    \caption{Cumulative proportion of label responses in the human recognition experiment as a function of semantic distance of the labels from the ground-truth label averaged over different concepts and abstraction levels.}
    \label{fig:SNP}
\end{figure}

In order to compute the semantic neighbor preference (SNP) score we first extracted semantic embeddings for each of the 1,854 THINGS concepts, which were previously computed by Hebart et al. (2023) based on human triplet similarity ratings. Next, for each concept we computed the rank of remaining 1,853 concepts in terms of their proximity to to the target concept.
Determining the SNP score for sketches of a given concept at a given level of abstraction involved obtaining the ranks of all the top-1 guesses of all the sketches of that type and excluding sketches where the guess matched the ground-truth. This was because the SNP score measures how semantically related \textit{off-target} labels are to the true label. We next computed the cumulative proportion of labels provided for that kind of sketch as a function of rank. Thus, if most guesses were semantic neighbors of the true label then the cumulative proportion would increase and reach a value of 1 quickly. In contrast, if many of the guesses were semantically distant from the true label, their ranks would be much lower and thus the cumulative proportion would also increase slowly.
Fig. \ref{fig:SNP} shows how this cumulative proportion of labels varies as a function of rank (normalized from 1-1853 to 0-1) averaged over all sketch types. The SNP score is the area under the curve (AUC) in this plot such that an AUC value closer to 1 indicates that guesses are generally in the same semantic neighbourhood as the ground-truth label, an AUC value of 0.5 indicates that labels are random and an AUC value close to 0 indicates that guesses are systematically far away from the ground truth. 

\subsubsection{Details On Fitting Linear Regression Models}

Standard ordinary least squares regression models were fit using the \texttt{statsmodels} package in \texttt{Python}.

Mixed-effects linear regression models reported in this paper were fit in \texttt{R} using the \texttt{lme4} package version 1.1.30.
We generally made use of the \texttt{bobyqa} optimizer when fitting mixed-effects models.

\textbf{Humans produce sparser sketches under stronger time constraints}. We fit models predicting the number of strokes from the drawing duration (42, 82, 16s, or 32s). We included random intercepts and slopes for the effect of drawing time for each concept.

\textbf{Sparser sketches are more semantically ambiguous for models and humans}
We fit separate sets of models for predicting human and vision model performance.
For human performance we fit 3 regression models predicting top-1 classification accuracy, response entropy, and semantic neighbor preference with drawing duration as the only predictor in each. We included random intercepts and slopes for the effect of drawing time for each concept.

For vision model performance we also fit 3 separate models predicting top-1 classification accuracy, response entropy, and semantic neighbor preference with drawing duration and model type as predictors. Model-type was a dummy-coded variable coding for each of the 17 vision models with CLIP as the reference level.

\textbf{Different models display distinct patterns of sketch recognition behavior}

To test whether top-1 classification accuracy, response entropy, and semantic neighbor preference varied across each of the 17 vision models, we also fit reduced versions of the models described in the earlier section \textit{without} a predictor for model-type. 
We next performed a $\chi^2$ model comparison between the full and reduced models to test whether the amount of variance explained by the regression models that did include the model-type predictor was significantly different from the regression models that did not include the predictor.

\textbf{A large gap remains between human and model sketch understanding}

We estimated human-baselines by splitting the human data in half multiple times and fitting linear regression models predicting top-1 classification accuracy, response entropy, and semantic neighbor preference values of one half of the data from the other half with an interaction term for draw duration to account for variance within each `abstraction level'. We report the mean of the adjusted model $R^2$ values across multiple splits of the data as the baseline values. 

When computing human-model alignment we also fit linear regression models predicting vision-model top-1 classification accuracy, response entropy, and semantic neighbor preference values from the human values for the corresponding sketches.
Here too, we included an interaction term for draw duration in each regression model. We report adjusted $R^2$ as a measure of human-model alignment.
To compute confidence intervals we sampled the dataset of all sketches multiple times with replacement, generating bootstrapped samples of the same size as the original dataset.
We bootstrapped samples within each concept and draw duration level to ensure that each sample was balanced in terms of the number of sketches of each concept at each level of abstraction.

\textbf{A CLIP-based sketch generation algorithm emulates human sketches under some conditions}

In order to compare human top-1 classification of human-made and \texttt{CLIPasso}-generated sketches, we fit a linear regression model predicting human top-1 classification accuracy from \texttt{CLIPasso} classification accuracy of the same type of sketches. We included an interaction term for abstraction level, similar to earlier analyses. We report the adjusted $R^2$ value of the model as a measure of how comparably recognizable human and \texttt{CLIPasso} sketches were.

\textit{Human-CLIPasso response divergence}. To measure the degree to which human label responses diverged between human-made and \texttt{CLIPasso}-generated sketches of the same concept at the same level of abstraction, we adopted the following procedure. For each type of machine sketch (e.g., lion made with 4 strokes), we computed the Jensen-Shannon distance (JSD) between that type of machine sketch and \textit{all} types of human sketches (4s lions, 8s lions, $\ldots$ 16s tanks, 32s tanks). Next, we ranked all the human sketches in terms of their proximity to the machine sketch and recorded the minimum rank for human sketches belonging to the same concept. For example, if the machine sketch was a 4-stroke lion and the ranks of the 4s, 8s, 16s, and 32s human lion sketches were 7,13,39, and 71 respectively then we would record \textbf{7}, the rank corresponding to the 4s human sketch.
This number is our reported measure of human-CLIPasso response divergence. In essence, our measure answers the question - Do \textit{any} human lion sketches elicit similar labels as a 4-stroke \texttt{CLIPasso} lion sketch?